\documentclass{article}

\usepackage{amsmath,amssymb,amsthm}
\usepackage{hyperref}
\usepackage{multirow}
\usepackage{graphicx}
\usepackage{subfigure}
\usepackage{wrapfig}
\usepackage{bbm}
\usepackage{epstopdf}
\usepackage[utf8]{inputenc}

\theoremstyle{plain} \newtheorem{thm}{Theorem}
 \newtheorem{lem}[thm]{Lemma}
\newtheorem{prop}[thm]{Proposition}

\theoremstyle{definition} \newtheorem{defn}{Definition}

\theoremstyle{remark} 

  \newenvironment{Proof}{\noindent{\bf Proof} \ }{\QED}\smallskip
\newcommand\QED{\newline \rightline{$\blacksquare$} \bigskip}

  \smallskip

  \newenvironment{Proof7}{\noindent{\bf Proof of Theorem 7} \ }{\QED}\smallskip

  \newenvironment{Proof thm4}{\noindent{\bf Proof of Theorem 4} \ }{\QED}\smallskip

  \newenvironment{Proof thm5}{\noindent{\bf Proof of Theorem 5} \ }{\QED}\smallskip

  \newenvironment{Proof thm6}{\noindent{\bf Proof of Theorem 6} \ }{\QED}\smallskip

\newcommand\Rp{\mathcal{R}_{L,P,\lambda}}
\newcommand\RT{\mathcal{R}_{L,T}}
\newcommand\RP{\mathcal{R}_{L,P}}

\newcommand\h{\mathcal{H}}
\newcommand\e{\epsilon_n}
\newcommand\X{\mathcal{X}}
\newcommand\x{\mathbf{X}}
\newcommand\F{\mathcal{F}}

\newcommand{\R}{\mathbf{R}}
\newcommand{\p}{\mathbf{P}}
\newcommand{\E}{\mathbf{E}}

\title{Learning Minimum Volume Sets and Anomaly Detectors from KNN Graphs}
\author{Jonathan Root, Venkatesh Saligrama\footnote{This paper was presented in part at NIPS 2009 and AISTATS 2015.}, Jing Qian}
\begin{document}

\maketitle

\begin{abstract}
We propose a non-parametric anomaly detection algorithm for high dimensional
data. We first rank scores derived from nearest neighbor graphs on $n$-point
nominal training data. We then train limited complexity models to imitate these
scores based on the max-margin learning-to-rank framework. A test-point is declared as an anomaly
at $\alpha$-false  alarm level if the predicted score is in the $\alpha$-percentile.
The resulting anomaly detector is shown to be asymptotically optimal in that for
any false alarm rate $\alpha$, its decision region converges to the $\alpha$-percentile
minimum volume level set of the unknown underlying density. In addition, we test
both the statistical performance and computational efficiency of our
algorithm on a number of synthetic and real-data experiments. Our results
demonstrate the superiority of our algorithm over existing $K$-NN based
anomaly detection algorithms, with significant computational savings.
\end{abstract}

\section{Introduction}\label{sec:intro}

Anomaly detection involves detecting statistically significant deviations of test
data from expected behavior. Such expected behavior is 
characterized by a nominal distribution. In typical applications the nominal
distribution is unknown and generally cannot be reliably
estimated from nominal training data due to a combination of factors
such as limited data size and high dimensionality.

We propose an adaptive non-parametric method for anomaly detection
based on a score function mapping the data set into the interval 
$[0,1]$. Our score function is derived from a $K$-nearest neighbor
graph (K-NNG), which orders the $n$-point nominal data set according
to their individual $K$-nearest neighbor distances. 
Anomaly is declared
whenever the score of a test sample falls below $\alpha$
(the desired ``false alarm'' error). The efficacy of our method
rests upon its close connection to multivariate $p$-values. In
statistical hypothesis testing, a $p$-value is any transformation
of the feature space to the interval $[0,1]$ that induces a 
uniform distribution on the nominal data. When test samples
with $p$-values smaller than $\alpha$ are declared as anomalous,
the false alarm error is less than $\alpha$. 

We develop a novel notion of $p$-values based on measures of level sets
of likelihood ratio functions. Our notion provides a characterization of the
optimal anomaly detector. More specifically, it is uniformly the most powerful
test for a specified false alarm level for the case when the 
anomaly density is a mixture of the nominal and a known 
density. We show that our score function is asymptotically
consistent, namely, it converges to the multivariate $p$-value
as data length approaches infinity. Motivated by this approach,
in aim of computational savings, we train limited complexity
models to imitate these scores based on the max-margin
learning to-rank framework. We prove consistency in
this setting as well.

Anomaly detection has been extensively studied. It is also referred
to as novelty detection \cite{ref:LP_novelty_detection,ref:AD_survey_markou1}, outlier detection \cite{ref:Ramaswamy2000}, one-class classification
\cite{ref:consistency_vert06} and single-class classification \cite{ref:single_class_Yaniv07} in the literature. Approaches to 
anomaly detection can be grouped into several categories. In 
parametric approaches \cite{ref:para_1993} the nominal densities are assumed
to come from a parametrized family and generalized likelihood 
ratio tests are used for detecting deviations from nominal. It is 
difficult to use parametric approaches when the distribution is 
unknown and data is limited. A $K$-nearest neighbor (K-NN)
anomaly detection approach is presented in \cite{ref:Ramaswamy2000,ref:knn_2009}. There an 
anomaly is declared whenever the distance to the $K$-th 
nearest neighbor of the test sample falls outside a threshold.
In comparison our anomaly detector utilizes the global  
information available from the entire K-NN graph to 
detect deviations from the nominal. In addition it has
provable optimality properties.  Learning theoretic approaches
attempt to find decision regions, based on nominal data, that
separate nominal instances from their outliers. These include
one-class SVM of Scholkopf et. al. \cite{ref:Scholkopf2000}
where the basic idea is to map the training data into the kernel
space and to separate them from the origin with maximum margin.
While these approaches provide impressive computationally efficient
solutions on real data, it is generally difficult to precisely relate tuning
parameter choices to desired false alarm probability.

Scott and Nowak \cite{ref:MV_2006} derive decision regions based on
minimum volume (MV) sets, which does provide Type I and Type II 
error control. They approximate (in appropriate function classes) level
sets of the unknown nominal density from training samples. Related
work by Hero \cite{ref:GEM_2006} based on geometric entropic minimization
(GEM) detects outliers by comparing test samples to the most concentrated
subset of points in the training sample. This most concentrated set is
the $K$-point minimum spanning tree (MST) for $n$-point nominal 
data and converges asymptotically to the minimum entropy set
(which is also the MV set). Nevertheless, computing $K$-MST 
for $n$-point data is generally intractable. To overcome these 
computational limitations \cite{ref:GEM_2006} proposes heuristic
greedy algorithms based on leave-one out K-NN graphs, which
while inspired by the $K$-MST algorithm is no longer provably optimal.
Our approach is related to these latter techniques, namely, MV sets
of \cite{ref:MV_2006} and GEM approach of \cite{ref:GEM_2006}. 
We develop score functions on K-NNG which turn out to be
the empirical estimates of the volume of the MV sets containing
the test point. The volume, of course a real number, is a sufficient
statistic for ensuring optimal guarantees. In this way we avoid
explicit high-dimensional level set computation. Yet our algorithm 
leads to statistically optimal solutions with the ability to control
false alarm and miss error probabilities. This paper extends
our preliminary work \cite{aistats15} by developing a more 
systematic and in depth approach. 

The rest of the paper is organized as follows. In section 2 we introduce
the problem setting and motivation. In section 3 we define the $p$-value
and provide a brief explanation as to why and how it is used to
derive the uniformly most powerful test for anomaly detection. 
Section 4 is devoted to score functions which imitate the 
$p$-value, and a proof of consistency. In section 5 
we show how we imitate these scores while preserving
consistency. We give a proof of 
the consistency of our algorithm is given and a 
finite sample bound is derived. In section 6 we
present our algorithm in detail and
synthetic and real experiments are 
also reported.

\section{Problem Setting \& Motivation}

Let $S=\lbrace x_1, x_2, ..., x_n\rbrace $ be a given set of nominal $d$-dimensional data points in the unit cube $[0,1]^d$.
We assume each data point $x_i$ to be sampled i.i.d from an unknown density $f_0$ supported on $[0,1]^d$. The problem is to assume a new data point,
$\eta\in \R^d$, is given, and test whether $\eta$ follows the distribution
of $S$. If $f$ denotes the density of this new (random) data point, then the set-up is summarized in the following hypothesis test:
\[
H_0: f=f_0 \;\;\;\;\; \text{vs.} \;\;\;\;\; H_1: f\neq f_0.
\]
We look for a functional $D:\R^d\to \R$ such that $D(\eta)>0\implies$ $\eta$ nominal.
Given such a $D$, we define its corresponding acceptance region by $A= \{ x : D(x)>0\}$.
We will see below that $D$ can be defined by the $p$-value.

Given a prescribed significance level (false alarm level) $\alpha\in (0,1)$, we require the probability that $\eta$ {\it does not} deviate from the nominal ($\eta \in A$), given $H_0$, to be bounded below by $1-\alpha$.
We denote this distribution by $\p_0$  :
\[
\p_0(A)=\p(\text{not } H_1 | H_0)=\int_A f_0(x) \; dx \geq 1-\alpha.
\]
Said another way, the probability that $\eta$ {\it does} deviate from the nominal,
given $H_0$, should fall under the specified significance level $\alpha$:
\[
1-\p_0(A)=\p( H_1 | H_0) \leq \alpha.
\]
At the same time we would like to minimize the probability of
predicting $\eta$ to be nominal, when in fact it is anomalous. This
is described by the event $\eta \in A$, given $H_1$, with 
probability:
 \[
 \int_A f(x) \; dx .
 \]
 This is sometimes known as the false negative. The complement
 of this event is then to be maximized, and this is known
 as the detection power:
 \[
 1-\int_A f(x) \; dx.
 \]
 
We assume $f$ to be bounded
above by a constant $C$, in which case $\int_A f(x) \; dx \leq C\cdot \lambda(A)$,
where $\lambda$ is Lebesgue measure on $\R^d$. The problem of finding
the most suitable acceptance region, $A$, can therefore
be formulated as finding the following minimum volume set:
\begin{equation}\label{MV}
U_{1-\alpha}:=\arg\min_{A} \left\{ \lambda(A) : \int_A f_0(x) \; dx \geq 1-\alpha \right\}.
\end{equation}
In words, we seek a set $A$ which captures at least a fraction $1-\alpha$ of
the probability mass, of minimum volume. 
In this case our decision rule, $D(\eta)>0\implies$
$\eta$ nominal, is said to the be the uniformly most
powerful test at the prescribed significance level $\alpha$.

\section{The $p$-value}
Assuming the existence of a functional $D:\R^d\to \R$,
we have shown how the problem of anomaly detection
can be formulated as one of finding a minimum volume
set. We now describe the desired functional, namely
the $p$-value. 

The set-up is the same as above, except now we specify
the test point $\eta$ to come from a mixture distribution,
namely $f(\eta)= (1-\pi)f_0(\eta)+\pi f_1(\eta)$, where
$f_1$ is a mixing density supported on $[0,1]^d$. 

\begin{defn}
Given a measure space $(X, \mu)$, and a measurable 
function $f:X\to \R$, we say that $f$ has no {\it non-zero
flat spots} on $X$ if for any $x\in X$ and
$\sigma>0$,
\[
\mu\{ y: |f(y)-f(x)|<\sigma\} < M\sigma,
\]
for some constant $M$.
\end{defn}

\begin{defn}
Let $\p_0$ be the nominal probability measure and $f_1$
a $\p_0$ -measurable function. Suppose the likelihood
ratio $f_1(x)/f_0(x)$ has no non-zero flat spots on any
open ball in $[0,1]^d$. Then we define the {\it $p$-value}
of a point $\eta\in[0,1]^d$ as
\[
p(\eta):= \p_0\left( x : \frac{f_1(x)}{f_0(x)}\geq \frac{f_1(\eta)}{f_0(\eta)}\right).
\]
\end{defn}

The $p$-value is a $\p_0$ - measurable map, and the 
distribution of $p(\eta)$ under $H_0$ is uniform on
$[0,1]$. To build intuition about the transformation and
its utility, consider the following example. When the mixing
density is uniform, namely, $f_1(\eta)=U(\eta)$ where
$U(\eta)$ is the uniform density over $[0,1]^d$, note that
$\Omega_{\alpha}=\{\eta : p(\eta) \geq \alpha\}$ is a density
level set at level $\alpha$: the collection of $\eta$ such that
\[
   p(\eta) = \p_0\left( x : f_0(x)\leq f_0(\eta)\right)= \int_{\{x : f_0(x)\leq f_0(\eta)\}}f_0(x)\; dx \geq \alpha.
 \]
In this case, it is not hard to see that 
\[
\p_0\{ \eta : p(\eta)\geq \alpha \}  = 1-\alpha,
\]
confirming that $p(\eta)$ is indeed uniformly distributed 
under $H_0$. It follows that for a given significance level $\alpha$,
the functional $D(\eta):=p(\eta)- \alpha$ defines the minimum
volume set in (\ref{MV}):
  \[
  U_{1-\alpha} = \{x : p(x)\geq \alpha\}.
  \]
The generalization to arbitrary $f_1$ is described next. 

\begin{thm}
The uniformly most powerful test for testing $H_0 : \pi =0$ versus
the alternative (anomaly) $H_1 : \pi >0$ at a prescribed level $\alpha$
of significance $\p_0( H_1 : H_0) \leq \alpha$ is 
\[
\phi(\eta) = \begin{cases} H_1, & p(\eta) \leq \alpha \\
                                         H_0, & otherwise.
                                         \end{cases}
\]
\end{thm}

\begin{Proof}
We provide the main idea for the proof. First, measure theoretic arguments
are used to establish that $p(X)$ is a random variable over $[0,1]$ under
both nominal and anomalous distributions. Next when $X\sim f_0$,
the random variable $p(X) \sim U([0,1])$. When $X \sim f = (1-\pi)f_0+
\pi f_1$ with $\pi>0$ we have $p(X) \sim g$, where $g$ is a 
monotonically decreasing PDF supported on $[0,1]$. Consequently,
the uniformly most powerful test for a significance level $\alpha$ is to
declare $p$-values smaller than $\alpha$ as anomalies.
\end{Proof}

 \section{Score Functions Based on K-NNG
 and Consistency} \label{sec:scorefunc}
 
 \subsection{Score Functions}
We have shown, assuming technical conditions on the density $f_0$, that
the $p$-value defines the minimum volume set:
  \[
  U_{1-\alpha} = \{x : p(x)\geq \alpha\}.
  \]
  Thus if we know $p$, we know the minimum volume set,
  and we can declare anomaly simply by checking whether or not
  $p(\eta) < \alpha$. However, $p$
  is based on information from the unknown density $f_0$, hence we
  must estimate $p$.

  Set $d(x,y)$ to be the Euclidean metric on $\R^d$.
  Given a point $x\in \R^d$,  we form its associated $K$ nearest
  neighbor graph (K-NNG), relative to $S$,
  by connecting it to the $K$ closest points in $S\setminus \{x\}$. 
  Let $R_S(x)$ denote the distance from $x$ to its $K$th nearest neighbor
  in $S\setminus \{x\}$.  We also define the related quantity,
  $N_S(x)$, which is the number of points in $S\setminus\{x\}$
  within a distance  $\epsilon$ of $x$. Said another way,
  $N_S(x)$ counts the number of data points from $S$
  within the ball of radius $\epsilon$, centered at $x$, not
  including $x$. This quantity is associated with
  the $\epsilon$ nearest neighbor graph ($\epsilon$-NNG)
  which connects all points from the data set to its 
  neighbors within a distance of $\epsilon$. 
  
  Associated to these two notions, we define
  two score functions:
  \begin{equation}\label{estimate_p}
  \hat{p}_K(\eta) := \frac{1}{n} \sum_{i=1}^n \mathbb{I}_{\{ R_S(\eta) \leq R_S(x_i)\}}
  \end{equation}
  and
  \begin{equation}\label{epsilon}
  \hat{p}_{\epsilon}(\eta) = \frac{1}{n} \sum_{i=1}^n 
   \mathbb{I}_{\{ N_S(\eta) \geq N_S(x_i)\}}
\end{equation}
  These functions measure the concentration of the point 
  $\eta$ relative to the training set. Intuitively, the larger the
  value of the density $f_0(\eta)$ at the point $\eta$, the {\it smaller}
  we expect $R_S(\eta)$ to be; analogously, the larger the value of
  the density $f_0(\eta)$ at the point $\eta$, the {\it larger} we expect $N_S(\eta)$
  to be. Assuming then that $f_1$ is uniform, one expects 
  \begin{equation}\label{con}
\lim_{n\to \infty}   \hat{p}_{\epsilon}(\eta) = p(\eta) \;\;\; \text{a.s.}
\end{equation}
and similarly for $\hat{p}_{K}(\eta)$:
\begin{equation}\label{con'}
\lim_{n\to \infty}   \hat{p}_{K}(\eta) = p(\eta) \;\;\; \text{a.s.}
\end{equation}
We develop the proof of these claims now.

\subsection{Theory: A Proof of Consistency}

To begin, we may assume $\eta \in [0,1]^d$ is fixed. Indeed, $\eta$
is drawn independent of $S$, hence we may as well condition
on $\eta$ and obtain (\ref{con}) and (\ref{con'}), then undo the 
conditioning. Such arguments will be used freely in what follows.

As defined, $R_S(\eta)$ and $R_S(x_i)$ are correlated because the 
neighborhoods of $\eta$ and $x_i$ might overlap. To overcome this
difficulty, we split the data set in two. Assume $n=2m+1$ (say), and
divide $S$ as 
\[
S=S_1\cup S_2 = \{x_0,x_1,\dots, x_m\} \cup \{ x_{m+1},\dots, x_{2m}\}.
\]
We modify our two score functions as 
\[
\hat{p}_{\epsilon}(\eta) = \frac{1}{m} \sum_{x_i\in S_1}  \mathbb{I}_{\{ N_{S_2}(\eta) \geq N_{S_1}(x_i)\}}
\]
and
\[  
 \hat{p}_{K}(\eta) = \frac{1}{m} \sum_{x_i\in S_1}  \mathbb{I}_{\{ R_{S_2}(\eta) \leq R_{S_1}(x_i)\}} .
 \]
Now $R_{S_2}(\eta)$ and $R_{S_1}(x_i)$ are independent.   

We will also require the density $f_0$ to satisfy the following 
regularity conditions:
\begin{itemize}
\item
$f_0$ is $C^2$ with $\lVert \nabla f_0(x) \rVert \leq \lambda$, and
\item
the Hessian matrix $H(x)$ of $f_0(x)$ is always dominated by a matrix
with largest eigenvalue $\lambda_M$. 
\end{itemize}

We organize our result in the following theorem:

\begin{thm}\label{KNN_Consistency}
Consider the set-up above with training data $S=\{x_1,\dots, x_n\}$
generated i.i.d. from $f_0$. Let $\eta \in [0,1]^d$ be an arbitrary
test sample. It follows that for a suitable choice of $K=K_n$, under
the above regularity conditions, 
\[
\lim_{n\to \infty}   \hat{p}_{K}(\eta) = p(\eta) \;\;\; \text{a.s.}
\]
\end{thm}

For the proof of this theorem we proceed in three steps:
\begin{enumerate}
\item
We show that
the expectation $\E_{S_1,S_2}[\hat{p}_{\epsilon}(\eta)]
\to p(\eta)$. 
\item
This result is then extended to $\hat{p}_K(\eta)$.
\item
Finally we show that $\hat{p}_K(\eta)$ concentrates about its mean,
$\E_{S_1,S_2}[\hat{p}_K(\eta)]$. 
\end{enumerate}

This is the content of the next three lemmas.

\begin{lem} Let $S=S_1\cup S_2$ as above. We have 
\[
\E_{S_1,S_2}[\hat{p}_{\epsilon}(\eta)]= \E_{x_1}[\p^{x_1}_{S_1,S_2}(
N_{S_2}(\eta) \geq N_{S_1}(x_1))]
\]
where $\p^{x_1}_{S_1,S_2}$ denotes the probability over $S_1,S_2$,
conditioned on $x_1$. 
Moreover, 
\[
\ell_m(\eta,x_1) \leq \p^{x_1}_{S_1,S_2}(N_{S_2}(\eta) \geq N_{S_1}(x_1))
\leq u_m(\eta,x_1)
\]
where both $\ell_m(\eta,x_1)$ and $u_m(\eta,x_1)$ converge to 
$\mathbb{I}_{\{f_0(\eta)\geq f_0(x_1)\}}$ as $m\to \infty$, and thus
\[
\E_{S_1,S_2}[\hat{p}_{\epsilon}(\eta)]\to p(\eta).
\]
\end{lem}

\begin{Proof}

By interchanging the expectation with the summation,
\begin{eqnarray*}
\E_{S_1,S_2}\left[\hat{p}_{\epsilon}(\eta)\right]
&=&\E_{S_1,S_2}\left[\frac{1}{m}\sum_{x_i\in S_1}
 \mathbb{I}_{\{N_{S_2}(\eta)\geq N_{S_1}(x_i)\}}\right]\\
&=&\frac{1}{m}\sum_{x_i\in S_1}
\E_{S_1,S_2}\left[
 \mathbb{I}_{\{N_{S_2}(\eta)\geq N_{S_1}(x_i)\}}\right]\\
&=&\E_{S_1,S_2}\left[\mathbb{I}_{\{N_{S_2}(\eta)\geq N_{S_1}(x_1)\}}\right]\\
&=& \E_{x_1}[\p_{S_1,S_2}^{x_1}(N_{S_2}(\eta)\geq N_{S_1}(x_1))].
\end{eqnarray*}
In the third line we have used that $N_{S_2}(\eta)-N_{S_1}(x_i)$ is of equal 
distribution for each $x_i\in S_1$, and in the fourth line we have
denoted the conditional probability with respect to $x_1$ by
$\p^{x_1}$. 

We must show that
\[
\p_{S_1,S_2}^{x_1}(N_{S_2}(\eta)\geq N_{S_1}(x_1)) \to \mathbb{I}_{\{
f_0(\eta) \geq f_0(x_1)\}}.
\]
To prove this, first condition on $S_2$ and
note that $N_{S_1}(x_1)$ is a binomial 
random variable with success probability
\begin{equation}\label{success}
q(x_1) = \int_{B_\epsilon} f_0(x_1+t) \, dt
\end{equation}
on $m$ trials. Hence by the Chernoff bound,
\[
\p_{S_1}^{x_1}(N_{S_1}(x_1) - mq(x_1) \geq \delta) \leq 
 \exp\left(\frac{-\delta^2}{2mq(x_1)}\right).
 \]
This implies,
\begin{equation}\label{eqn_chernoff}
\p_{S_1}^{x_1,S_2}( N_{S_2}(\eta) \geq N_{S_1}(x_1) )
\geq  \mathbb{I}_{\{N_{S_2}(\eta) \geq mq(x_1)+\delta_{x_1}\}} - 
   \exp\left(\frac{-\delta_{x_1}^2}{2mq(x_1)}\right).
\end{equation}
We choose $\delta_{x_1} = q(x_1) m^{\gamma}$ ($\gamma$ will
be specified later), and reformulate equation (\ref{eqn_chernoff})
as
\begin{equation}\label{chernoff}
\p_{S_1}^{x_1,S_2}( N_{S_2}(\eta) \geq N_{S_1}(x_1) )
\geq  \mathbb{I}_{\big\{\frac{N_{S_2}(\eta)}{m\text{Vol}(B_{\epsilon})} \geq \frac{q(x_1)}{\text{Vol}(B_{\epsilon})}\left(1+\frac{1}{m^{1-\gamma}}\right)\big\}} - 
   \exp\left(\frac{-q(x_1)m^{2\gamma-1}}{2}\right).
   \end{equation}
To incorporate $f_0(x_1)$ into this equation, we use the 
smoothness conditions to approximate (\ref{success}) 
by its Taylor approximation:
\begin{eqnarray*}
\left| \frac{ \int_{B_\epsilon} f_0(x_1+t) \, dt}{\text{Vol}(B_{\epsilon})}
-f_0(x_1) \right|   &\leq& \frac{\lambda}{\text{Vol}(B_{\epsilon})}\int_{B_\epsilon}\lVert t\rVert\, dt +\frac{1}{2}\frac{\lambda_M}{\text{Vol}(B_\epsilon)}\int_{B_\epsilon} \lVert t\rVert^2 \, dt \\
                           &=&
                \frac{d}{d+1}\lambda \epsilon + \frac{d}{2(d+2)}\lambda_M \epsilon^2.
\end{eqnarray*}
With this equation (\ref{chernoff}) becomes,
\[
\p_{S_1}^{x_1,S_2}( N_{S_2}(\eta) \geq N_{S_1}(x_1) )
\geq  \mathbb{I}_{\big\{\frac{N_{S_2}(\eta)}{m\text{Vol}(B_{\epsilon})} \geq \left( f_0(x_1) +\frac{d}{d+1}\lambda \epsilon + \frac{d}{2(d+2)}\lambda_M \epsilon^2\right)
\left(1+\frac{1}{m^{1-\gamma}}\right)\big\}} - 
  \exp\left(\frac{-q(x_1)m^{2\gamma-1}}{2}\right).
   \]
Taking expected values of both sides (with respect to $S_2$),
we obtain
\begin{equation}\label{eqn_eta}
\p_{S_1,S_2}^{x_1}( N_{S_2}(\eta) \geq N_{S_1}(x_1) ) \geq
\end{equation}
\[
  \p_{S_2}\left(\frac{N_{S_2}(\eta)}{m\text{Vol}(B_{\epsilon})} \geq \left( f_0(x_1) +\frac{d}{d+1}\lambda \epsilon + \frac{d}{2(d+2)}\lambda_M \epsilon^2\right)
\left(1+\frac{1}{m^{1-\gamma}}\right)\right) - 
  \exp\left(\frac{-q(x_1)m^{2\gamma-1}}{2}\right).
   \]
   Applying the Chernoff bound to $N_{S_2}(\eta)$, with $\delta_{\eta}
   =q(\eta)m^{\gamma}$,
  \begin{equation}\label{chernoff_eta}
  \p_{S_2}\left( \frac{N_{S_2}(\eta)}{m\text{Vol}(B_{\epsilon})}
  \geq \frac{q(\eta)}{\text{Vol}(B_{\epsilon})}\left(1-\frac{1}{m^{1-\gamma}}\right)
  \right) \geq 1- \exp\left(\frac{-q(\eta)m^{2\gamma-1}}{2}\right).
  \end{equation}
  Using the Taylor expansion in (\ref{chernoff_eta}),
  \begin{equation}\label{taylorchernoff}
  \p_{S_2}\left(\frac{N_{S_2}(\eta)}{m\text{Vol}(B_{\epsilon})} \geq \left( f_0(\eta) -\frac{d}{d+1}\lambda \epsilon - \frac{d}{2(d+2)}\lambda_M \epsilon^2\right)
\left(1-\frac{1}{m^{1-\gamma}}\right)\right) \geq 
1- \exp\left(\frac{-q(\eta)m^{2\gamma-1}}{2}\right).
\end{equation}
It follows that ,
\begin{equation}\label{morechernoff}
\p_{S_1,S_2}^{x_1}( N_{S_2}(\eta) \geq N_{S_1}(x_1) ) \geq
\end{equation}
\[
  \mathbb{I}_{\big\{\left(f_0(\eta) -\frac{d}{d+1}\lambda \epsilon - \frac{d}{2(d+2)}\lambda_M \epsilon^2\right)\left(1-\frac{1}{m^{1-\gamma}}\right) \geq \left( f_0(x_1) +\frac{d}{d+1}\lambda \epsilon + \frac{d}{2(d+2)}\lambda_M \epsilon^2\right)
\left(1+\frac{1}{m^{1-\gamma}}\right)\big\}} 
\]
\[
- 
\exp\left(\frac{-q(\eta)m^{2\gamma-1}}{2}\right) -
  \exp\left(\frac{-q(x_1)m^{2\gamma-1}}{2}\right) =: \ell_m(\eta, x_1).
   \]
Indeed, if we have the inequality 
\[
\left(f_0(\eta) -\frac{d}{d+1}\lambda \epsilon - \frac{d}{2(d+2)}\lambda_M \epsilon^2\right)\left(1-\frac{1}{m^{1-\gamma}}\right) \geq 
\]
\[
\left( f_0(x_1) +\frac{d}{d+1}\lambda \epsilon + \frac{d}{2(d+2)}\lambda_M \epsilon^2\right)
\left(1+\frac{1}{m^{1-\gamma}}\right)
\]
then from (\ref{taylorchernoff}), we have with probability (with
respect to $\p_{S_2}$) at least
$1- \exp\left(\frac{-q(\eta)m^{2\gamma-1}}{2}\right)$, 
\[
\frac{N_{S_2}(\eta)}{m\text{Vol}(B_{\epsilon})} \geq \left( f_0(x_1) +\frac{d}{d+1}\lambda \epsilon + \frac{d}{2(d+2)}\lambda_M \epsilon^2\right)
\left(1+\frac{1}{m^{1-\gamma}}\right).
\]
Using this in (\ref{eqn_eta}), (\ref{morechernoff}) follows. 

A similar upper bound follows analogously. Briefly, if $N_{S_1}(x_1)\geq
\delta_{\eta}+mq(\eta)$, then
\[
\p_{S_2}^{x_1,S_1}(N_{S_2}(\eta) \geq N_{S_1}(x_1)) \leq
\exp\left( \frac{-\delta_{\eta}^2}{2mq(\eta)}\right).
\]
It follows that 
\begin{eqnarray*}
\p_{S_2}^{x_1,S_1} (N_{S_2}(\eta) \geq N_{S_1}(x_1)) &\leq& 
\mathbb{I}_{\{N_{S_1}(x_1) \leq \delta_{\eta} +mq(\eta)\}} + 
\exp\left( \frac{-\delta_{\eta}^2}{2mq(\eta)}\right)\\
                &=&  \mathbb{I}_{\big\{\frac{N_{S_1}(x_1)}{m\text{Vol}(B_{\epsilon})} \leq \frac{q(\eta)}{\text{Vol}(B_{\epsilon})}\left(1+\frac{1}{m^{1-\gamma}}\right)\big\}} +
   \exp\left(\frac{-q(\eta)m^{2\gamma-1}}{2}\right) \\
                &\leq & \mathbb{I}_{\big\{\frac{N_{S_1}(x_1)}{m\text{Vol}(B_{\epsilon})} \leq\left( f_0(\eta) +\frac{d}{d+1}\lambda \epsilon + \frac{d}{2(d+2)}\lambda_M \epsilon^2\right)\left(1+\frac{1}{m^{1-\gamma}}\right)\big\}} +
   \exp\left(\frac{-q(\eta)m^{2\gamma-1}}{2}\right).
   \end{eqnarray*}
 Taking expected values with respect to $S_1$, and applying the Chernoff/
 Taylor bound              
   \begin{equation*}
  \p_{S_1}\left(\frac{N_{S_1}(x_1)}{m\text{Vol}(B_{\epsilon})} \leq \left( f_0(x_1) -\frac{d}{d+1}\lambda \epsilon - \frac{d}{2(d+2)}\lambda_M \epsilon^2\right)
\left(1-\frac{1}{m^{1-\gamma}}\right)\right) \leq 
 \exp\left(\frac{-q(x_1)m^{2\gamma-1}}{2}\right),
\end{equation*}
we obtain
\begin{equation}\label{upper}
\p_{S_1,S_2}^{x_1}( N_{S_2}(\eta) \geq N_{S_1}(x_1) ) \leq
\end{equation}
\[
  \mathbb{I}_{\big\{\left(f_0(\eta) +\frac{d}{d+1}\lambda \epsilon + \frac{d}{2(d+2)}\lambda_M \epsilon^2\right)\left(1+\frac{1}{m^{1-\gamma}}\right) \geq \left( f_0(x_1) -\frac{d}{d+1}\lambda \epsilon - \frac{d}{2(d+2)}\lambda_M \epsilon^2\right)
\left(1-\frac{1}{m^{1-\gamma}}\right)\big\}} 
\]
\[
+ 
\exp\left(\frac{-q(\eta)m^{2\gamma-1}}{2}\right) +
  \exp\left(\frac{-q(x_1)m^{2\gamma-1}}{2}\right) =: u_m(\eta, x_1).
   \]
(\ref{upper}) should be read: if we have the inequality 
\[
\left(f_0(\eta) +\frac{d}{d+1}\lambda \epsilon + \frac{d}{2(d+2)}\lambda_M \epsilon^2\right)\left(1+\frac{1}{m^{1-\gamma}}\right) \leq \left( f_0(x_1) -\frac{d}{d+1}\lambda \epsilon - \frac{d}{2(d+2)}\lambda_M \epsilon^2\right)
\left(1-\frac{1}{m^{1-\gamma}}\right)
\]
then 
\[
\p_{S_1,S_2}^{x_1}( N_{S_2}(\eta) \geq N_{S_1}(x_1) ) \leq
\exp\left(\frac{-q(\eta)m^{2\gamma-1}}{2}\right) +
  \exp\left(\frac{-q(x_1)m^{2\gamma-1}}{2}\right).
  \]
 
  Putting (\ref{morechernoff}) and (\ref{upper}) together, we have
  \[
\ell_m(\eta,x_1)\leq \p^{x_1}_{S_1,S_2}(N_{S_2}(\eta)\geq N_{S_1}(x_1))
 \leq u_m(\eta,x_1)
\]
Choosing $\epsilon=\epsilon_m\to 0$ in $m$, and $\gamma=5/6$,
we obtain 
\[
\p_{S_1,S_2}^{x_1}(N_{S_2}(\eta)\geq N_{S_1}(x_1)) \to \mathbb{I}_{\{
f_0(\eta) \geq f_0(x_1)\}}
\]
as $m\to \infty$. By the dominated convergence theorem, 
$\E_{S_1,S_2}[\hat{p}_{\epsilon}(\eta)]=\E_{x_1}[
\p_{S_1,S_2}^{x_1}(N_{S_2}(\eta)\geq N_{S_1}(x_1))]
\to p(\eta)$, as desired. 
\end{Proof}

We now extend this result to $\hat{p}_K(\eta)$. The objective
is to show that 
 \[
\p_{S_1,S_2}^{x_1}(R_{S_2}(\eta)\leq R_{S_1}(x_1)) \to \mathbb{I}_{\{
f_0(\eta) \geq f_0(x_1)\}}.
\]

\begin{lem}\label{Knn}
Set $K= \frac{\pi^{d/2}}{\Gamma(d/2+1)}(1-2m^{-1/6})m^{2/5}(f_0(\eta)-\Delta_1)$,
with $\Delta_1=\frac{d}{d+1}\lambda \epsilon + \frac{d}{2(d+2)}\lambda_M\epsilon^2$,
and $\epsilon=m^{-\frac{3}{5d}}$. Then with this choice of $K=K_m$ and 
$\epsilon=\epsilon_m$, we have 
\[
\p^{x_1}_{S_1,S_2}(R_{S_2}(\eta)\leq R_{S_1}(x_1)) \to \mathbb{I}_{\{f_0(\eta) > f_0(x_1)\}}.
\]
\end{lem}

\begin{Proof}
Let $K=K_m$, $\epsilon=\epsilon_m$ 
and consider the event 
$\{N_{S_2}(\eta)\geq K\} \cap \{N_{S_1}(x_1)\leq K\}$,
or equivalently
\[
\{N_{S_2}(\eta) - q(\eta)m\geq K-q(\eta)m\} 
\cap \{ N_{S_1}(x_1) -q(x_1)m \leq K-q(x_1)m\}.
\]
Using the Chernoff bound, the probability of the above two
events both converge to one exponentially fast if
\[
K-q(\eta)m < 0 \;\;\;\;\; \text{and} \;\;\;\;\; K-q(x_1)m>0.
\]
By the Taylor approximation,
\[
\frac{q(\eta)}{\text{Vol}(B_{\epsilon})}\leq f_0(\eta)+\Delta_1
\]
\[
\frac{q(\eta)}{\text{Vol}(B_{\epsilon})}\geq f_0(\eta)-\Delta_1
\]
where $\Delta_1=\frac{d}{d+1}\lambda \epsilon + \frac{d}{2(d+2)}\lambda_M\epsilon^2$.
So it will suffice if
\[
K-m\text{Vol}(B_{\epsilon})(f_0(\eta)-\Delta_1)<0
\]
\[
K-m\text{Vol}(B_{\epsilon})(f_0(x_1)+\Delta_1)>0
\]
where Vol$(B_{\epsilon})= \frac{\pi^{d/2}}{\Gamma(d/2+1)} \epsilon^d$.
Choosing $\epsilon= m^{-\frac{3}{5d}}$ (say), this is satisfied if
\[
K-  \frac{\pi^{d/2}}{\Gamma(d/2+1)} m^{2/5}(f_0(\eta)-\Delta_1)<0
\]
\[
K-  \frac{\pi^{d/2}}{\Gamma(d/2+1)} m^{2/5}(f_0(x_1)+\Delta_1)>0
\]
Set $K= \frac{\pi^{d/2}}{\Gamma(d/2+1)}(1-2m^{-1/6})m^{2/5}(f_0(\eta)-\Delta_1)$
and $\Delta_2= 2m^{-1/6}$. Then the first condition on $K$ is
satisfied, and for the second we must have
\[
(1-\Delta_2)(f_0(\eta) - \Delta_1) > f_0(x_1)+\Delta_1.
\]

Putting this all together, first note that for any
$K=K_m$ and $\epsilon=\epsilon_m$, 
\begin{eqnarray*}
\{N_{S_2}(\eta) \geq K_m\}_{\epsilon_m} \cap \{N_{S_1}(x_1) \leq K_m\}_{\epsilon_m} &=&
\{R_{S_2}(\eta) \leq \epsilon_m\}_{K_m} \cap \{R_{S_1}(x_1) \geq \epsilon_m\}
_{K_m}  \\
   &\subseteq&  \{R_{S_2}(\eta) \leq R_{S_1}(x_1)\}_{K_m},
   \end{eqnarray*}
   where we have used the subscript notation in
   $\{N_{S_2}(\eta) \geq K_m\}_{\epsilon_m} \cap \{N_{S_1}(x_1) \leq K_m\}_{\epsilon_m}$, for example, to carefully indicate that this event depends
   on the radius, $\epsilon_m$, of the ball within which the $\epsilon_m$
   neighbors are being counted. Similarly, the event 
   $\{R_{S_2}(\eta) \leq \epsilon_m\}_{K_m} $ is dependent on $K_m$
   being specified. In some sense then, these two variables are inversely
   related (once a choice of $K$ and $\epsilon$ has been specified).
With our choice of $\epsilon=\epsilon_m$ and $K=K_m$, we therefore
have,
\begin{eqnarray*}
\p_{S_1,S_2}^{x_1}(R_{S_2}(\eta) \leq R_{S_1}(x_1)) &\geq& \p_{S_1,S_2}^{x_1}
(\{N_{S_2}(\eta) \geq K\} \cap \{N_{S_1}(x_1) \leq K\}) \\
     &\geq& \mathbb{I}_{\{K-q(\eta)m<0, K-q(x_1)m>0\}} -\exp(w_m)\\
     &\geq& \mathbb{I}_{\{(1-\Delta_2)(f_0(\eta)-\Delta_1) >f_0(x_1) +\Delta_1\}}-\exp(w_m)\\
     &\to& \mathbb{I}_{\{f_0(\eta) >f_0(x_1)\}}
     \end{eqnarray*}
where $\exp(w_m)\to 0$ as $m\to \infty$. 

It remains to show that for $f_0(\eta)<f_0(x_1)$, 
\[
\p_{S_1,S_2}^{x_1}(R_{S_2}(\eta) \leq R_{S_1}(x_1)) \to 0.
\]
The proof is by contradiction. If not, then there exists
$\epsilon_m$ and $K_m$ such that 
\[
\p_{S_1,S_2}^{x_1}(\{R_{S_2}(\eta) \leq \epsilon_m\}_{K_m} \cap \{R_{S_1}(x_1) \geq \epsilon_m\}_{K_m} ) \to c>0.
\]
Or equivalently,
\[
\p_{S_1,S_2}^{x_1}(\{N_{S_2}(\eta) \geq K_m\}_{\epsilon_m} \cap \{N_{S_1}(x_1) \leq K_m\}_{\epsilon_m} ) \to c>0
\]
This contradicts the result from Part 1, since 
$\{N_{S_2}(\eta) \geq K_m\}_{\epsilon_m} \cap \{N_{S_1}(x_1) \leq K_m\}_{\epsilon_m}\subseteq \{N_{S_2}(\eta) \geq N_{S_1}(x_1) \}_{\epsilon_m}$,
and we know the probability of the later event converges to zero for
$f_0(\eta)<f_0(x_1)$. 
\end{Proof}

We now verify that
$\hat{p}_K(\eta)$ satisfies the requirements of McDiarmid's 
inequality, i.e. has bounded differences. Set 
$F(x_1,\dots, x_{m}) = \hat{p}_K(\eta) = \frac{1}{m} \sum_{x_i\in S_1}
\mathbb{I}_{\{R_{S_2}(\eta) \leq R_{S_1}(x_i)\}}$. Using 
corollary 11.1 in \cite{ref:devroye} we have 
\[
|F(x_1,\dots, x_i,\dots, x_m) - F(x_1,\dots, x_i',\dots, x_m)| \leq \frac{K\gamma_d}{m}
\]
where $\gamma_d$ is a constant and is defined as the minimal number of 
cones centered at the origin of angle $\pi/6$ that cover $\R^d$. We have 
thus shown the following lemma:

\begin{lem}\label{simpleconcentration}
With $K=cm^{2/5}$ we have 
\[
\p_{S_1,S_2}(| \E_{S_1,S_2}[\hat{p}_K(\eta)] - \hat{p}_K(\eta) | >\delta)
\leq 2e^{-\frac{2\delta^2m^{1/5}}{c^2\gamma_d^2}}.
\]
\end{lem}

Theorem \ref{KNN_Consistency} now follows from the combination
of Lemma \ref{Knn} and Lemma \ref{simpleconcentration} and a 
standard application of the first Borel-Cantelli lemma.


\section{Score Functions Imitating K-NNG and Consistency}\label{Alteration}

\subsection{Altering the Score Function}

The consistency result of Theorem \ref{KNN_Consistency}
 is attractive from a statistical viewpoint, however the test-time complexity of the
$K$-NN distance statistic grows as $O(dn)$. 
This can be prohibitive for real-time applications. Thus we are compelled to
learn a score function respecting the $K$-NN distance statistic, but with
significant computational savings. 
This is achieved by mapping the data set $S$ into a reproducing kernel Hilbert
space (RKHS), $H$, with kernel $k$ and inner product $\langle \cdot, \cdot\rangle$.
We denote by $\Phi$ the mapping $\R^d\to H$,
defined by $\Phi(x_i) =k(x_i,\cdot)$. We then optimally learn a function $\hat{g}\in H$
respecting the ordering
\[
\{(i,j) : R_S(x_i) > R_S(x_j)\}
\]
and construct the scoring function as
\begin{equation}\label{estimate_pn}
\hat{R}_n(\eta) := \frac{1}{n} \sum_{i=1}^n \mathbb{I}_{\{\langle \hat{g},\Phi(\eta)\rangle < \langle \hat{g}, \Phi(x_i)\rangle\}}.
\end{equation}
It will turn out that $\hat{R}_n$ is an asymptotic estimator of the $p$-value 
and thus we will say a test point $\eta$ is anomalous if $\hat{R}_n(\eta)\leq \alpha$.

\begin{thm}\label{thm:rank_main_consistency}
With $K=O(n^{2/5})$, 
as $n\rightarrow\infty$, $\hat{R}_n(\eta)\rightarrow p(\eta)$  a.s.
\end{thm}

The difficulty in this theorem arises from the fact that the score function
$\hat{R}_n(\eta)$
is based on the KNN-distance statistic $R_S$, which is learned from data with high-dimensional
noise. Moreover, the noise is distributed according to an {\it unknown} probability
measure.  

For the proof of this theorem, we begin with the law of large numbers.
Suppose a function $G$ is found such that
  $f_0(x_i)<f_0(x_j)\iff G(x_i)<G(x_j)$ for all $x_i,x_j$
  in the data set $\{x_1,\dots, x_n\}$ as $n\to \infty$. 
  By our assumptions on $f_0$, given a test point $\eta$,
  there exists a point $x_i$ in the nominal data set such 
  that $d(x_i,\eta)$ is arbitrarily small. Thus we have
  the almost sure equality 
  \[
  \{f_0(x_1) < f_0(\eta)\} = \{ G(x_1) < G(\eta)\}.
  \]
  Therefore by the law of large numbers
\[
\frac{1}{n} \sum_{i=1}^n \mathbb{I}_{\{G(x_i) < G(\eta)\}} \rightarrow 
\E_0(\mathbb{I}_{\{G(x_1) < G(\eta) \}} )=p(\eta).
\]
Of course the KNN-distance statistic reverses the ordering of 
the density, in which case 
\[
\frac{1}{n} \sum_{i=1}^n \mathbb{I}_{\{G(x_i) > G(\eta)\}} \rightarrow 
\E_0(\mathbb{I}_{\{G(x_1) > G(\eta) \}} )=p(\eta).
\]
We must therefore show that $\hat{g}$ respects the ordering
of the density $f_0$. 

We begin by proving that $\hat{g}$ is consistent in the sense of \cite{ref:Steinwart2001}.
 Fix an RKHS $H$ on the input
space $X\subset \R^d$ with RBF kernel $k$. 
We denote by $L$ the hinge loss. We may
write $\hat{g}$ as the solution to the following regularized minimization 
problem,
\begin{equation}\label{regular}
\hat{g} = \arg \min_{f\in H} \mathcal{R}_{L,T}(f) +\lambda_n\lVert f \rVert_H^2,
\end{equation}
where 
\[
\mathcal{R}_{L,T}(f)=\frac{1}{n^2}\sum_{i,j}L(f(x_i)-f(x_j)).
\]
$T$ denotes the pairs from the sample $S=\{x_1,\dots,x_n\}$, so this is a loss with respect
to the empirical measure. The expected risk is denoted 
\[
\mathcal{R}_{L,P}(f) = E_{S}[\mathcal{R}_{L,T}(f)].
\]
Then consistency means that, under appropriate conditions as
$\lambda_n\to 0$ and $n\to \infty$ , we have 
\begin{equation}\label{consistent}
\E_{S}[ \mathcal{R}_{L,T}(\hat{g})] \to \min_{f\in H} \mathcal{R}_{L,P}(f).
\end{equation}
The proof of this claim requires a concentration of measure result
relating $\mathcal{R}_{L,T}(f)$ to its expectation, $\mathcal{R}_{L,P}(f)$,
uniformly over $f\in H$. The argument follows closely that made in
\cite{ref:Smale2001}.

Finally it is shown that if $\hat{g}$ satisfies (\ref{consistent}), then 
it ranks samples according to their density: $f_0(x_i)>f_0(x_j)\iff
\hat{g}(x_i)>\hat{g}(x_j)$. This is proposition \ref{thm_surrogate_condition}, and
 finishes the proof. 
 
 From proposition \ref{thm_surrogate_condition} we also deduce a
 global concentration inequality. Suppose we alter the data 
 set $S=\{x_1,\dots, x_n\}$ at one point to $S^i=\{x_1,\dots, x_i',\dots,x_n\}$.
 Then the solution $\hat{g}$ to the regularized minimization  problem (\ref{regular})
 with $S$ will rank the data according to the density; and the solution
 $\hat{g}_i$ of (\ref{regular}) with $S^i$ will rank the data $S^i$ according to the density. Thus
 the two estimators
 \[
\hat{R}_n(\eta) := \frac{1}{n} \sum_{j=1}^n \mathbb{I}_{\{\langle \hat{g},\Phi(\eta)\rangle < \langle \hat{g}, \Phi(x_i)\rangle\}}.
\]
and 
\[
\hat{R}_{n,i}(\eta) := \frac{1}{n} \sum_{j=1}^n \mathbb{I}_{\{\langle \hat{g}_i,\Phi(\eta)\rangle < \langle \hat{g}_i, \Phi(x_i)\rangle\}}.
\]
will differ in at most one summand, and thus by at most 1/n. This is a uniform estimate,
so 
 \[
F(x_1,\dots, x_n):=\hat{R}_n(\eta) 
\]
has bounded differences with constant $1/n$. Moreover, as
already noted, proposition \ref{thm_surrogate_condition} also
implies that $\E_S[\hat{R}_n(\eta)] =p(\eta)$. 
We now conclude from McDiarmid's inequality the following theorem.
\begin{thm}
With $K=O(n^{2/5})$ we have
\[
\p_S(|\hat{R}_n(\eta)-p(\eta)|\geq \epsilon) \leq e^{-2n\epsilon^2}.
\]
\end{thm}

\subsection{Proof of Theorem  \ref{thm:rank_main_consistency} }\label{sec:analysis}
We fix an RKHS $H$ on the input space $X\subset\R^d$
with an RBF kernel $k$. We abstract the set-up as follows. 
Let $S=\{x_1,\dots,x_n\}$ be a set of objects to be ranked
in $\R^d$ with labels 
$\mathbf{r}= \{r_1,\dots,r_n\}$ (e.g. $R_S(x_i)=r_i$).
Here $r_i$ denotes the label of $x_i$, and $r_i\in \R$.
We assume the variables in $S$ to be distributed according
to $P$, and $\mathbf{r}$ deterministic.   Throughout
$L$ denotes the hinge loss.

The following notation will be useful in the proof of Theorem \ref{thm:rank_main_consistency}. 
Take $T$ to be the set of pairs derived from $S$ and
define the $L$-$risk$ of $f\in H$ as
\[
\RP (f) := \E_{S}[\RT(f)]
\]
where
\[
\RT(f)=\sum_{i,j:r_i>r_j}D(r_i,r_j)L(f(x_i)-f(x_j))
\]
and $D(r_i,r_j)$ is some positive weight function, which we take for simplicity to be
$1/|\mathcal{P}|$, $\mathcal{P} = \{(i,j) : r_i>r_j\}$. $\RT(f)$ is the {\it empirical} $L$-risk
of $f$, with respect to the empirical distribution over the pairs of samples.
The smallest possible $L$-risk in $H$ is denoted
\[
\RP:= \inf_{f\in H} \RP(f).
\]
The {\it regularized} $L$-$risk$ is
\begin{equation}\label{regularize}
\Rp^{\text{reg}}(f):=\lambda \lVert f\rVert^2+\mathcal{R}_{L,P}(f),
\end{equation}
 $\lambda >0$.

For simplicity we assume the preference pair set $\mathcal{P}$ contains all pairs over these $n$ samples.
Let $g_{S,\lambda}$  be the optimal solution to our algorithm.
We have,
\begin{equation}\label{eq_emp_solution}
  g_{S,\lambda} = \arg \min_{f\in H} \RT(f) + \lambda || f ||^2
\end{equation}

Let $\mathcal{H}_n$ denote a ball of radius $O(1/\sqrt{\lambda_n})$
in $H$. Let $C_k:= \sup_{x,t}|k(x,t)|$ with $k$ the rbf kernel associated
to $H$. Given $\epsilon>0$, we let
$N(\mathcal{H},\epsilon/4C_k)$ be the covering number
of $\mathcal{H}$ by disks of radius $\epsilon/4C_k$ .
We first show that with appropriately chosen $\lambda$, as $n\rightarrow\infty$,
$g_{S,\lambda}$ is consistent in the following sense.

\begin{prop}\label{thm_convergence_surrogate}
Let $\lambda_n$ be appropriately chosen such that $\lambda_n\rightarrow 0$ and $\frac{\log N(\h_n,\epsilon/4C_k)}{n\lambda_n} \to 0$, as $n\rightarrow\infty$. Then we have
\[
    \E_{S}[ \RT(g_{S,\lambda_n})] \rightarrow  \RP=\min_{f\in H}\RP(f), \;\;\; n\to \infty.
\]
\end{prop}
\begin{Proof}
Let us outline the argument. In \cite{ref:Steinwart2001}, the author shows that there exists a $f_{P,\lambda}\in H$
minimizing (\ref{regularize}):
\begin{itemize}
\item
For all Borel probability measures $P$ on $X\times X$ and
all $\lambda >0$, there is an $f_{P,\lambda} \in H$ with
\[
\Rp^{\text{reg}}(f_{P,\lambda}) = \inf_{f\in H} \Rp^{\text{reg}}(f)
\]
such that $\lVert f_{P,\lambda} \rVert =O(1/\sqrt{\lambda})$.
(If $P$ is the empirical distribution over data $T$, then we denote this minimizer
by $f_{T,\lambda}$.)
\end{itemize}

Next, a simple argument shows that
\begin{itemize}
\item
$\lim_{\lambda\to 0} \Rp^{\text{reg}}(f_{P,\lambda})= \RP.$
\end{itemize}

Finally, we will need a concentration inequality to relate the $L$-risk
of $f_{P,\lambda}$ with the empirical $L$-risk of $f_{P,\lambda}$.
We then derive consistency using the following argument:
\begin{align*}
\RP(f_{T,\lambda_n})
& \leq  \lambda_n \lVert f_{T,\lambda_n}\rVert^2+\mathcal{R}_{L,P}(f_{T,\lambda_n})\\
& \leq \lambda_n \lVert f_{T,\lambda_n}\rVert^2+\mathcal{R}_{L,T}(f_{T,\lambda_n})+\delta/3  \\
& \leq  \lambda_n \lVert f_{P,\lambda_n}\rVert^2+\mathcal{R}_{L,T}(f_{P,\lambda_n})+\delta/3 \\
& \leq  \lambda_n \lVert f_{P,\lambda_n}\rVert^2+\mathcal{R}_{L,P}(f_{P,\lambda_n})+2\delta/3  \\
& \leq  \mathcal{R}_{L,P}+\delta
\end{align*}
where $\lambda_n$ is an appropriately chosen sequence $\to 0$,
and $n$ is large enough. The second and fourth inequality hold due to Concentration Inequalities, and the last one holds since $\lim_{\lambda \to 0} \Rp^{\text{reg}}(f_{P,\lambda})=\mathcal{R}_{L,P}$.

We now prove the appropriate concentration inequality \cite{ref:Smale2001}.
Recall $H$ is an RKHS with
smooth kernel $k$; thus the inclusion $I_{k}: H\to C(X)$ is compact, where
$C(X)$ is given the $\lVert \cdot \rVert_{\infty}$-topology. That is, the
``hypothesis space'' $\mathcal{H}:= \overline{I_k(B_R)}$ is compact in $C(X)$,
where $B_R$ denotes the ball of radius $R$ in $H$. We denote by $N(\mathcal{H},
\epsilon)$ the covering number of $\mathcal{H}$ with disks of radius $\epsilon$.
We prove the following inequality:

\begin{lem}
For any probability distribution $P$ on $X\times X$,
\begin{equation*}
P^{\e}\{T\in (X\times X)^{\e}:\sup_{f\in \h} | \RT(f)-\RP(f)| \geq \epsilon \}\leq
2N(\h,\epsilon/4C_k)\exp\left(
\frac{-\epsilon^2n}{2(1+2\sqrt{C_k}R)^2}\right),
\end{equation*}

where $C_k := \sup_{x,t}|k(x,t)|$.
\end{lem}

\begin{Proof}
Since $\h$ is compact, it has a finite covering number. Now suppose $\h=
D_1\cup \cdots \cup D_{\ell}$ is any finite covering of $\h$. Then
\begin{equation*}
P\{\sup_{f\in \h} | \RT(f)-\RP(f)| \geq \epsilon \} \leq  \sum_{j=1}^{\ell}
P\{\sup_{f\in D_j} | \RT(f)-\RP(f)| \geq \epsilon \}
\end{equation*}
so we restrict attention to one of the disks $D_j$ in $\h$ of radius $\epsilon$.

Suppose $\lVert f-g \rVert_{\infty}\leq \epsilon$. We want to show that the
difference
\[
|(\RT(f)-\RP(f))-(\RT(g)-\RP(g))|
\]
is also small. Rewrite this quantity as
\[
 |(\RT(f)-\RT(g))-E_{S}[\RT(g)-\RT(f)]|.
 \]
 Since $\lVert f-g \rVert_{\infty}\leq \epsilon$, for $\epsilon$ small enough we have
\begin{align*}
\max\{0,1-(f(x_i)-f(x_j))\}-\max\{0,1-(g(x_i)-g(x_j))\} & = \max\{0,(g(x_i)-g(x_j)-f(x_i)+f(x_j))\} \\
& = \text{max}\{0,\langle g-f, \phi(x_i)-\phi(x_j)\rangle\}.
\end{align*}
Here $\phi:X\to H$ is the feature map, $\phi(x):=k(x,\cdot)$.
 Combining this with the Cauchy-Schwarz inequality, we have
\begin{eqnarray*}
 |(\RT(f)-\RT(g))-E_{\bold{x}}[\RT(g)-\RT(f)]| & \leq  \frac{2}{n^2}(2n^2\lVert f-g \rVert_{\infty}C_k) & \leq 4C_k\epsilon,
\end{eqnarray*}
where $C_k:= \sup_{x,t} |k(x,t)|$. From this inequality it follows that
\begin{equation*}
|\RT(f)-\RP(f)| \geq (4C_k+1)\epsilon  \implies |(\RT(g)-\RP(g))| \geq \epsilon.
\end{equation*}
We thus choose to cover $\h$ with disks of radius $\epsilon/4C_k$, centered at
$f_1,\dots,f_{\ell}$. Here $\ell= N(\h,\epsilon/4C_k)$ is the covering number
for this particular radius. We then have
\begin{equation*}
\sup_{f\in D_j}|(\RT(f)-\RP(f))|\geq 2\epsilon  \implies |(\RT(f_j)-\RP(f_j))|\geq \epsilon.
\end{equation*}
Therefore,
\begin{equation*}
 P\{\sup_{f\in \h} | \RT(f)-\RP(f)| \geq 2\epsilon \}  \leq
 \sum_{j=1}^n P\{ | \RT(f_j)-\RP(f_j)| \geq \epsilon \}
 \end{equation*}
The probabilities on the RHS can be bounded using McDiarmid's inequality.

Define the random variable $g(x_1,\dots,x_n) :=\mathcal{R}_{L,T}(f)$, for fixed $f\in H$.
We need to verify that $g$ has bounded differences. If we change one of the
variables, $x_i$, in $g$ to $x_i'$, then at most $n$ summands will change:
\begin{align*}
|g(x_1,\dots,x_i,\dots,x_n)-g(x_1,\dots,x_i',\dots,x_n)|
& \leq \frac{1}{n^2}2n\sup _{x,y} |1-(f(x)-f(y))| \\
& \leq \frac{2}{n}+\frac{2}{n}\sup_{x,y}|f(x)-f(y)|\\
& \leq \frac{2}{n}+\frac{4}{n}\sqrt{C_k}\lVert f \rVert.
\end{align*}
Using that $\sup_{f\in \mathcal{H}}\lVert f \rVert\leq R$,
McDiarmid's inequality thus gives
\begin{equation*}
P\{\sup_{f\in \h} | \RT(f)-\RP(f)| \geq \epsilon \}
 \leq 2N(\h,\epsilon/4C_k)\exp\left(
\frac{-\epsilon^2n}{2(1+2\sqrt{C_k}R)^2}\right).
\end{equation*}
\end{Proof}

We are now ready to prove Theorem 2. 
Take $R=\lVert f_{P,\lambda} \rVert$ and apply this result to $f_{P,\lambda}$:
\begin{equation*}
P\{| \RT(f_{P,\lambda})-\RP(f_{P,\lambda})| \geq \epsilon \}  \leq
2N(\h,\epsilon/4C_k)\exp\left(
\frac{-\epsilon^2n}{2(1+2\sqrt{C_k}\lVert f_{P,\lambda} \rVert)^2}\right).
\end{equation*}
Since $\lVert f_{P,\lambda_n} \rVert =O(1/\sqrt{\lambda_n})$, the RHS converges to 0
so long as $\dfrac{n\lambda_n}{\log N(\h,\epsilon/4C_k)} \to \infty$ as $n\to \infty$.
This completes the proof of Theorem 2. 
\end{Proof}

We now establish that under mild conditions on the surrogate loss function, the solution minimizing the expected surrogate loss will asymptotically recover the correct preference relationships given by the density $f$.
\begin{prop}\label{thm_surrogate_condition}
Let $L$ be a non-negative, non-increasing convex surrogate loss function
that is differentiable at zero and satisfies $L'(0)<0$. If
\begin{equation*}
  \hat{g} = \arg \min_{g\in H} E_{S} \left[ \RT(g) \right],
\end{equation*}
then $\hat{g}$ will correctly rank the samples according to their density, i.e.
$\forall x_i\neq x_j, f(x_i)> f(x_j) \implies \hat{g}(x_i)>\hat{g}(x_j)$.
\end{prop}


\begin{Proof} Our proof follows similar lines of Theorem 4 in \cite{ref:RDPS2012}.
Assume that $g(x_i) < g(x_j)$, and define a function $g'$ such that
$g'(x_i)=g(x_j)$, $g'(x_j)=g(x_i)$, and $g'(x_k)=g(x_k)$ for all $k\neq i,j$.
We have $\RP(g')-\RP(g)=E_{S}(A(S))$, where

\begin{eqnarray*}
 A(S)
   = \sum_{k : r_j<r_i<r_k}[D(r_k,r_j)-D(r_k,r_i)]  [L(g(x_k)-g(x_i))-L(g(x_k)-g(x_j))]
  \\ +  \sum_{k : r_j<r_k<r_i}D(r_i,r_k)[L(g(x_j)-g(x_k))-L(g(x_i)-g(x_k))]
  \\ +  \sum_{k : r_j<r_k<r_i}D(r_k,r_j)[L(g(x_k)-g(x_i))-L(g(x_k)-g(x_j))]
  \\ +  \sum_{k : r_j<r_i<r_k}[D(r_k,r_j)-D(r_k,r_i)][L(g(x_k)-g(x_i))-L(g(x_k)-g(x_j))]
  \\ +  \sum_{k : r_j<r_i<r_k}[D(r_i,r_k)-D(r_j,r_k)][L(g(x_j)-g(x_k))-L(g(x_i)-g(x_k))]
  \\ +  (L(g(x_j)-g(x_i))-L(g(x_i)-g(x_j)))D(r_i,r_j).
  \end{eqnarray*}
  Using the requirements of the weight function $D$ and the assumption that $L$
  is non-increasing and non-negative, we see that all six sums in the above
  equation for $A(\bold{x})$ are negative. Thus $A(S)<0$, so
  $\RP(g')-\RP(g)=E_{S}(A(S))<0$, contradicting the minimality
  of $g$. Therefore $g(x_i)\geq g(x_j)$.

  Now we assume that $g(x_i)=g(x_j)=g_0$. Since $\RP(g)=\inf_{h\in H}\RP(h)$,
  we have $\left. \dfrac{\partial{\ell_L(g;x)}}{\partial{g(x_i)}}\right|_{g_0}=A=0,$ and
  $\left. \dfrac{\partial{\ell_L(g;x)}}{\partial{g(x_j)}}\right|_{g_0}=B=0$, where
  \begin{eqnarray*}
  A=\sum_{k : r_j < r_i < r_k} D(r_k, r_i) [ -L'(g(x_k)-g_0)]+
    \sum_{k : r_j < r_k< r_i} D(r_i, r_k) L'(g_0-g(x_k)) +\\
  \sum_{k : r_k < r_j < r_i} D(r_i, r_k)  L'(g_0-g(x_k))+D(r_i,r_j)[-L'(0)].
  \end{eqnarray*}
  \begin{eqnarray*}
  B=\sum_{k : r_j < r_i < r_k} D(r_k, r_j) [ -L'(g(x_k)-g_0)]+
  \sum_{k : r_j < r_k< r_i} D(r_k, r_j) L'(g_0-g(x_k)) +\\
  \sum_{k : r_k < r_j < r_i} D(r_j, r_k)  L'(g_0-g(x_k))+D(r_i,r_j)[-L'(0)].
  \end{eqnarray*}
  However, using $L'(0)<0$ and the requirements of $D$ we have
  \[
  A-B\leq 2L'(0)D(r_i,r_j)<0,
  \]
  contradicting $A=B=0$.
\end{Proof}

\subsection{A Finite-Sample Generalization Result
and Minimum Volume Set Region Convergence}\label{subsec:finite_sample}

Recall that the anomaly detection problem was reformulated as a 
minimum volume set estimation problem in (\ref{MV}):
\[
U_{1-\alpha}=\arg\min_{A} \left\{ \lambda(A) : \int_A f_0(x) \; dx \geq 1-\alpha \right\},
\]
$\lambda$ denoting Lebesgue measure on $\R^d$. Moreover we concluded that
the so-called $p$-value -- based on the unknown nominal and anomalous probability
distributions -- defines the minimum volume set: 
\[
U_{1-\alpha} = \{ x : p(x) \geq \alpha\}.
\]
In our work in section \ref{Alteration} we construct a score function
$\hat{R}_n$ based on a sample $\{x_1,\dots,x_n\}$
of the nominal density $f_0$, and prove 
\begin{itemize}
\item
$\hat{R}_n(\eta) \to p(\eta)$ a.s.
\item
$\p_{0}( | \hat{R}_n(\eta) -p(\eta)| \geq \epsilon) \leq e^{-2n\epsilon^2}$.
\end{itemize}
Thus if $x\sim f_0$, the random set
\[
R_{\alpha} = \{ x: \hat{R}_n(x)\geq \alpha\}
\]
converges to $U_{1-\alpha}$ almost surely as the data size $n\to \infty$.
In this section we, consider the non-asymptotic estimate
of $\p_0(R_{\alpha})$, and study how well it satisfies
$\p_0(R_{\alpha}) = \int_{R_{\alpha}} f_0(x) \, dx
\geq 1-\alpha$.

Recall our notation. We learn a ranger $\hat{g}=g_{n}$, and compute the values $g_{n}(x_1),\ldots,g_{n}(x_n)$. Let
$g_n^{(1)}\leq g_n^{(2)}\leq \cdots \leq g_n^{(n)}$ be the ordered permutation of these values.
For a test point $\eta$, we evaluate $g_{n}(\eta)$ and compute $\hat{R}_n(\eta)$.
For a prescribed false alarm level $\alpha$, we define the decision region for claiming anomaly by
\begin{eqnarray*}
  R_\alpha^c &=& \{ x: \, \hat{R}_n(x) \leq \alpha  \} \\
   &=&  \{ x: \, \sum_{j=1}^{n} \textbf{1}_{\{ g_n(x)\leq g_n(x_j) \}} \leq \alpha n \}  \\
   &=&  \{ x: \, g_n(x) \geq g_n^{ (\lfloor n-\alpha n +1\rfloor )} \}.
\end{eqnarray*}

We give a finite-sample bound on the probability that a newly drawn nominal point $\eta$ lies in $R_\alpha$. In the following Theorem, $\mathcal{F}$ denotes a
real-valued function class of kernel based linear functions
equipped with the $\ell_{\infty}$ norm over
a finite sample $\mathbf{x}=\{x_1,\dots,x_n\}$:
\[
\lVert f \rVert _{\ell_{\infty}^{\mathbf{x}}} =\max_{x\in \mathbf{x}} |f(x)|.
\]
Note that $\mathcal{F}$ contain solutions
to an SVM-type problem, so we assume the output of our rankAD
algorithm, $g_n$, is an element of $\mathcal{F}$.
We let $\mathcal{N}(\gamma,\mathcal{F},n)$ denote the covering
number of $\mathcal{F}$ with respect to this norm.

\begin{thm}\label{thm:AD_finite_sample}
Fix a distribution $P$ on $\R^d$ and suppose $\x=\{x_1,\dots,x_n\}$ are generated iid from $P$.
For $g\in\mathcal{F}$ let $g^{(1)}\leq g^{(2)}\leq \cdots \leq g^{(n)}$ be the ordered permutation of
$g(x_1),\dots,g(x_n)$. Then with probability $1-\delta$ over such an $n$-sample,
for any $g\in \mathcal{F}$, $1\leq m \leq n$ and sufficiently small $\gamma>0$,
\begin{equation}\label{finitesample'}
 P\left\{x: g(x)< g^{(m)} - 2\gamma\right\} \leq \frac{m-1}{n} + \epsilon(n,k,\delta),
\end{equation}
where $\epsilon(n,k,\delta)= \frac{2}{n} (k+1+\log\frac{n}{\delta} ) $,
 $k=\lceil{\log\mathcal{N}(\gamma,\mathcal{F},2n)}\rceil$. 
 Similarly, with probability $1-\delta$ over an $n$-sample, for any $g\in \mathcal{F}$
 and small $\gamma>0$,
 \begin{equation}\label{finitesample}
P\left\{x: g(x)> g^{(m)} + 2\gamma\right\} \leq \frac{n-m}{n} + \epsilon(n,k,\delta)
\end{equation}
where $\epsilon = \frac{2}{n}(k+1 + \log\frac{n}{\delta})$.
\end{thm}

Take the second inequality (\ref{finitesample}), for example. 
Set $m=\lfloor n- \alpha n+1 \rfloor$. Then the left hand side is 
precisely the probability that a test point drawn from the nominal distribution has a score below the $\alpha$ percentile. We see that this probability is bounded from above by 
\[
\frac{\alpha n -1}{n}  + \epsilon(n,k,\delta)
\]
which converges to $\alpha$ as $n\to \infty$. 
This theorem is true irrespective of $\alpha$ and so we have shown that we can simultaneously approximate multiple level sets, and furthermore
minimum volume sets. We record this corollary
as a theorem:

\begin{thm}
Given a sample $\{x_1,\dots, x_n\}$ generated iid from an unknown
nominal distribution $\p_0$, let $\hat{g}=g_n$ be the solution to the rankAD 
minimization step (\ref{eq:ranksvm_standard}). 
Given $\gamma>0$ small, define the induced
decision region for claiming anomaly by 
\[
R_{\alpha,\gamma}^c
:=\{ x: \, g_n(x) \geq g_n^{ (\lfloor n-\alpha n +1\rfloor )}+2\gamma \}.
\]
Then with probability $1-\delta$ over any such 
$n$-sample, for any small $\gamma>0$, 
\[
\p_0\{x : x\in R_{\alpha,\gamma} \} \geq 1-\frac{\alpha n -1}{n}  - \epsilon(n,k,\delta)
\]
where $\epsilon = \frac{2}{n}(k+1 + \log\frac{n}{\delta})$. In particular,
in the limit, 
\[
\p_0\{x : x\in R_{\alpha,\gamma} \} \geq 1-\alpha, \;\; \text{as } n\to \infty.
\]
Moreover, since $\hat{R}_n(\eta)\to p(\eta)$ almost surely,
we conclude that $R_{\alpha,\gamma}$ converges to the 
minimum volume set (\ref{MV}) (with the proper adjustment 
for $\gamma$).
\end{thm}


For the proof of \ref{thm:AD_finite_sample} we need the following lemma \cite{ref:vapnik1979}:

\begin{lem}\label{vapnik}
Let $\X$ be a set and $S$ a system of sets in $\X$, and $P$ a probability
measure on $S$. For $\x\in \X^{n}$ and $A\in S$, define $\nu_{\x}(A):= |\x\cap A|/n$.
If $n>2/\epsilon$, then
\begin{equation*}
P^{n}\left\{ \x : \sup_{A\in S} |\nu_{\x}(A)- P(A)|>\epsilon\right\}  \leq
2P^{2n}\left\{ \x\x' : \sup_{A\in S} |\nu_{\x}(A)-\nu_{\x'}(A)|>\epsilon/2\right\}.
\end{equation*}
\end{lem}


\begin{Proof7}
Consider lemma \ref{vapnik} with
\[
A=\{ x : f(x) < f^{(m)} - 2\gamma \}.
\]
Then with $\gamma$ small enough 
\[
\nu_{\x}(A)= |\{x_j\in \x : f(x_j) < f^{(m)} - 2\gamma\}|/n=
 \frac{m-1}{n}.
 \]
Then
\[
P^{n}\left\{ \x : \sup_{A\in S} |P(A)-\nu_{\x}(A)|>\epsilon\right\} = P^{n}(J).
\]
where 
\begin{equation*}
J:=  \Biggl\{  \x \in \X^n : \exists f\in \F,  P(A) > \frac{m-1}{n} + \epsilon \Biggr\}
\end{equation*}
(Actually we have removed the absolute value, since lemma \ref{vapnik}
still holds.)
$J$ is the complement of 
the event (\ref{finitesample'}), so 
we must show that $P^n(J)\leq \delta$ for $\epsilon = \epsilon(n,k,\delta)$.
By lemma \ref{vapnik}, we have 
\[
P^n(J)\leq P^{2n} \Biggl\{ \x\x' :  \exists f\in \F, | \{ x_j' \in \x' : f(x_j')  < f^{(m)} - 2\gamma\}| > (m-1) + \epsilon n/2   \Biggr\}.
\]

Now consider a $\gamma_k$-cover $U$ of $\mathcal{F}$ with respect to the 
pseudo-metric $\ell_{\infty}^{\x\x'}$, where the existence of 
\[
\gamma_k = \min\{\gamma : N(\gamma, \mathcal{F}, 2\ell) \leq 2^k\}
\]
is shown in \cite{ref:oc_svm2001} lemma 10. Suppose for some $f\in \mathcal{F}$,
\[
|\{x_j' \in \x' : f(x_j') < f^{(m)}-2\gamma\}| > (m-1) + \epsilon n/2
\]
We can find $\hat{f}\in U$ with $\lVert f-\hat{f} \rVert_{\ell_{\infty}^{\x\x'}} \leq
\gamma_k\leq \gamma$ and so $f(x) < f^{(m)} -2\gamma$ implies
$\hat{f}(x) <\hat{f}^{(m)} $. Therefore,
\[
| \{ x_j' \in \x' : \hat{f}(x_j') < \hat{f}^{(m)}\}| > (m-1) + \epsilon n/2
\] 
This gives the upper bound
\[
P^n(J) \leq 2P^{2n}\biggr\{ \x\x' : \exists \hat{f} \in U, |\{x_j' \in \x' : \hat{f}(x_j')
< \hat{f}^{(m)} \}| > (m-1) + \epsilon n/2  \Biggr\}. 
\]
And since $U$ has at most $2^k$ elements, the union bound gives 
\[
P^n(J) \leq 2\cdot2^kP^{2n}\biggr\{ \x\x' :  |\{x_j' \in \x' : \hat{f}(x_j')
< \hat{f}^{(m)} \}| > (m-1) + \epsilon n/2 \Biggr\}. 
\]
The probability on the right can be bounded using a standard swapping
permutation argument. Specifically, let $\Gamma_{2n}$ denote the 
set of all permutations on $[2n]=\{1,\dots,2n\}$ which swap
some of the elements in the first half with the corresponding elements
in the second half. Then $|\Gamma_{2n}| = 2^n$ and
\[
P^{2n}\biggr\{ \x\x' :  |\{x_j' \in \x' : \hat{f}(x_j')
< \hat{f}^{(m)} \}| > (m-1) + \epsilon n/2  \Biggr\}\leq
\]
\[
\sup_{\x\x'}\left[  \frac{1}{2^n} \sum_{\sigma\in \Gamma_{2n}} \mathbb{I} \left(
|\{ x_j' \in \sigma(\x') : \hat{f}(x_j') < \hat{f}^{(m)}\}| > m-1+\epsilon n/2\right)\right]
\]
where $\sigma(\x')= \{x_{\sigma(1)}',\dots,x_{\sigma(n)}'\}$
and $\sigma$ is swapping to the data set $\x$, so for example
$x_{\sigma(1)}' =x_1$ or $x_1'$. Since $\x$ already contains
$m-1$ data points satisfying $\hat{f}(x_j) < \hat{f}^{(m)}$,
given any double sample $\x\x'$ the maximum
number of permutations which leave the event true is 
$2^{n-{\epsilon{n}/2}}$. Thus the ratio is $2^{-\epsilon n/2}$,
so 
\[
P^n(J) \leq 2\cdot2^{k}\cdot2^{-\epsilon n/2} = 2^{k+1-\epsilon n/2}.
\]
Setting the right hand side equal to $\delta/n$ and solving for $\epsilon$,
we get $\epsilon = \frac{2}{n} (k+1+\log\frac{n}{\delta} ) $.

The proof of the second inequality follows analogously, taking 
$A=\{ x : f(x) > f^{(m)} + 2\gamma \}$ and 
\begin{equation*}
J:=  \Biggl\{  \x \in \X^n : \exists f\in \F, P\{ x : f(x) > f^{(m)} + 2\gamma \}  > \frac{n-m}{n}+\epsilon \Biggr\}.
\end{equation*}
Then 
\[
P^n(J) \leq 2^{k+1-\epsilon n/2}.
\]
And so $\epsilon = \frac{2}{n}(k+1 + \log\frac{n}{\delta})$.
\end{Proof7}

\section{Rank-Based Anomaly Detection Algorithm}\label{sec:main_algo}
In this section we describe our main algorithm for anomaly detection, and discuss several of its properties and advantages. 

\subsection{Anomaly Detection Algorithm}

We present detailed steps of our rank-based anomaly detection algorithm as follows.

\noindent\rule[0.5ex]{\linewidth}{1pt}
\noindent\textbf{Algorithm 1: Ranking Based Anomaly Detection (rankAD)}

\noindent\rule[0.5ex]{\linewidth}{1pt}
\noindent\textbf{1. Input:}

\noindent Nominal training data $S={\lbrace x_1, x_2, ..., x_n \rbrace}$, desired false alarm level $\alpha$, and test point $\eta$.

\noindent\textbf{2. Training Stage:}

\noindent(a) Calculate $K$th nearest neighbor distances $R_S(x_i)$, and calculate
$\hat{p}_K(x_i)$ for each nominal sample $x_i$, using Eq.(\ref{estimate_p}).

\noindent(b) Quantize $\{\hat{p}_K(x_i),\,i=1,2,...,n\}$ uniformly into $m$ levels: $r_q(x_i) \in \lbrace 1,2,..., m \rbrace$. Generate preference pairs $(i, j)$ whenever their quantized levels are different: $r_q(x_i) > r_q(x_j)$.

\noindent(c) 
Set $\mathcal{P}=\{(i,j) : r_q(x_i)> r_q(x_j)\}$. Solve:
\begin{eqnarray}\label{eq:ranksvm_standard}
  \min_{g,\xi_{ij}}: \, && \,\, \frac{1}{2} ||g||^2 + C \sum_{(i,j)\in\mathcal{P}} \xi_{ij} \\
\nonumber
  s.t. \, &&  \,\, \langle g,\, \Phi(x_i)-\Phi(x_j) \rangle \geq 1 - \xi_{ij}, \,\,\,\, \forall (i,j)\in \mathcal{P}  \\
\nonumber
          &&  \,\,  \xi_{ij} \geq 0
\end{eqnarray}

\noindent(d) Let $\hat{g}$ denote the minimizer. Compute and sort: $\hat{g}(\cdot)=\langle \hat{g}, \Phi(\cdot)\rangle$ on $S={\lbrace x_1, x_2, ..., x_n \rbrace}$.

\noindent\textbf{3. Testing Stage:}

\noindent(a) Evaluate $\hat{g}(\eta)$ for test point $\eta$.

\noindent(b) Compute the score: $\hat{R}_n(\eta) = \frac{1}{n}\sum^n_{i=1} \textbf{1}_{\{ \hat{g}(\eta) < \hat{g}(x_i) \}}$. This can be done through a binary search over sorted $\{ \hat{g}(x_i),i=1,...,n \}$.

\noindent(c) Declare $\eta$ as anomalous if $\hat{R}_n(\eta) \leq \alpha$.

\noindent\rule[0.5ex]{\linewidth}{1pt}

\noindent{\bf Remark 1:}\\
The standard learning-to-rank setup \cite{ref:ranksvm} is to assume non-noisy input pairs.
Our algorithm is based on noisy inputs, where the noise is characterized by an unknown, high-dimensional distribution.
Yet we are still able to show the asymptotic consistency of the obtained ranker in Sec.\ref{sec:analysis}.

\noindent{\bf Remark 2:} \\
For the learning-to-rank step Eq.(\ref{eq:ranksvm_standard}), we equip the RKHS $H$ with the RBF kernel $k(x,x') = \exp\left( -\dfrac{\lVert x-x'\rVert^2}{\sigma^2} \right)$. The algorithm parameter $C$ and RBF kernel bandwidth $\sigma$ can be selected through cross validation, since this step is a supervised learning procedure based on input pairs. We use cross validation and adopt the weighted pairwise disagreement loss (WPDL) from \cite{ref:RDPS2012} for this purpose.

\noindent{\bf Remark 3:} \\
The number of quantization levels, $m$, impacts training complexity as well as performance.
When $m=n$, all $n \choose 2$ preference pairs are generated. This scenario has the highest training complexity. Furthermore, large $m$ tends to more closely follow rankings obtained from $K$-NN distances, which may or may not be desirable. $K$-NN distances can be noisy for small training data sizes.
While this raises the question of choosing $m$, we observe that setting $m$ to be $3\sim5$ works fairly well in practice. We fix $m=3$ in all of our experiments.
$m=2$ is insufficient to allow flexible false alarm control, as will be demonstrated next.

\noindent{\bf Remark 4:} \\
Let us mention the connection with ranking SVM. Ranking SVM is an algorithm
for the learning-to-rank problem, whose goal is to rank unseen objects based
on given training data and their corresponding orderings. Our novelty lies
in building a connection between learning-to-rank and anomaly detection:
\begin{enumerate}
\item
While there is no such natural ``input ordering'' in anomaly detection, we 
create this order on training samples through their $K$-NN scores.
\item
When we apply our detector on an unseen object it produces a score
that approximates the unseen object's $p$-value. We theoretically
justify this linkage, namely, our predictions fall in the right
quantile (Theorem \ref{thm:AD_finite_sample}). We also empirically
show test-stage computational benefits. 
\end{enumerate}


\subsection{False alarm control}
In this section we illustrate through a toy example how our learning method approximates minimum volume sets.
We consider how different levels of quantization impact level sets. We will show that for appropriately chosen quantization levels
our algorithm is able to simultaneously approximate multiple level sets. In Section~\ref{sec:analysis} we show that the normalized score Eq.(\ref{estimate_pn}), takes values in $[0,1]$, and converges to the $p$-value function. Therefore we get a handle on the false alarm rate. So null hypothesis can be rejected at different levels simply by thresholding $\hat{R}_n(\eta)$.

{\bf Toy Example:} \\
We present a simple example in Fig. 1
to demonstrate this point. The nominal density $ f \sim 0.5 \mathcal{N} \left(
\left[ 4; 1 \right],
0.5 I
 \right)
+
0.5 \mathcal{N} \left(
\left[ 4 ; -1 \right],
0.5 I
 \right)
$.
We first consider single-bit quantization ($m=2$) using RBF kernels ($\sigma=1.5$) trained with pairwise preferences between $p$-values above and below $3$\%. 
This yields a decision function $\hat{g}_2(\cdot)$. The standard way is to claim anomaly when $\hat{g}_2(x)<0$, corresponding to the outmost orange curve in (a). We then plot different level curves by varying $c>0$ for $\hat{g}_2(x)=c$, which appear to be scaled versions  of the orange curve. 
While this quantization appears to work reasonably for $\alpha$-level sets with $\alpha=3$\%, for a different desired $\alpha$-level, the algorithm would have to retrain with new preference pairs. On the other hand, we also train rankAD with $m=3$ (uniform quantization) and obtain the ranker $\hat{g}_3(\cdot)$. We then vary $c$ for $\hat{g}_3(x)=c$ to obtain various level curves shown in (b), all of which surprisingly approximate the corresponding density level sets well.
We notice a significant difference between the level sets generated with 3 quantization levels in comparison to those generated for two-level quantization.
In the appendix we show that $\hat{g}(x)$ asymptotically preserves the ordering of the density, and from this conclude that our score function $\hat{R}_n(x)$ approximates multiple density level sets ($p$-values). Also see Section \ref{sec:analysis} for a discussion of this.
However in our experiments it turns out that we just need $m=3$ quantization levels instead of $m=n$ ($n \choose 2$ pairs) to achieve flexible false alarm control and do not need any re-training.

\begin{figure*}[htb!]
\begin{centering}
\begin{minipage}[t]{.44\textwidth}
\includegraphics[width = 1\textwidth]{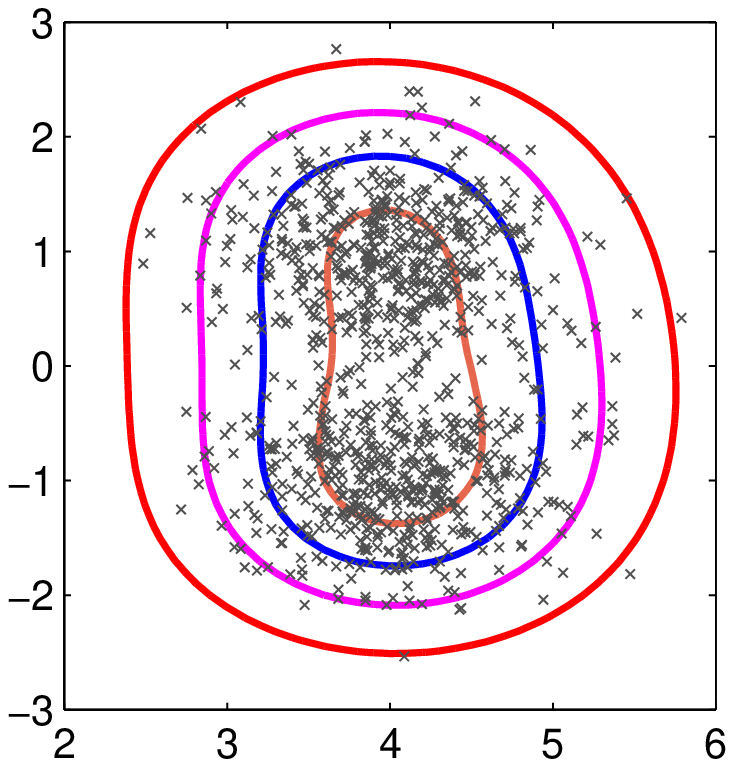}
\makebox[6 cm]{(a) Level curves ($m=2$)}\medskip
\end{minipage}
\begin{minipage}[t]{.44\textwidth}
\includegraphics[width = 1\textwidth]{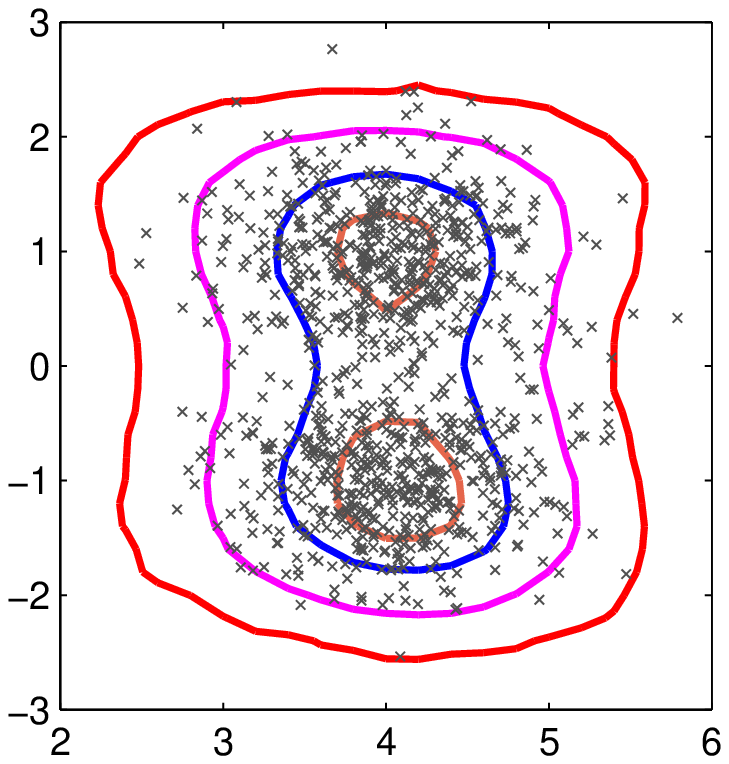}
\makebox[6 cm]{(b) Level curves ($m=3$)}\medskip
\end{minipage}
\caption{ Level curves of rankAD for different quantization levels. 1000 i.i.d. samples are drawn from a 2-component Gaussian mixture density. Left figure(a) depicts performance with single-bit quantization ($m=2$). To learn rankAD we quantized preference pairs at 3\% and $\sigma=1.5$ in our RBF kernel. Right figure(b) shows rankAD with 3-levels of quantization and $\sigma=1.5$.  (a) shows level curves obtained by varying the offset $c$ for $\hat{g}_2(x)=c$. Only the outmost curve ($c=0$) approximates the oracle density level set well while  the inner curves ($c>0$) appear to be scaled versions of outermost curve. (b) shows level curves obtained by varying $c$ for $\hat{g}_3(x)=c$. Interestingly we observe that the inner most curve approximates peaks of the mixture density. }
\end{centering}
\label{fig:level_curve}
\end{figure*}

\subsection{Time Complexity}
For training, the rank computation step requires computing all pair-wise distances among nominal points $O(dn^2)$, followed by sorting for each point $O(n^2\log n)$.
So the training stage has the total time complexity $O(n^2(d+\log n)+T)$, where $T$ denotes the time of the pair-wise learning-to-rank algorithm.
At test stage, our algorithm only evaluates $\hat{g}(\eta)$ on $\eta$ and does a binary search among $\hat{g}(x_1),\ldots,\hat{g}(x_n)$. The complexity is $O(ds +\log n)$, where $s$ is the number of support vectors. This has some similarities with one-class SVM where the complexity scales with the number of support vectors \cite{ref:oc_svm2001}. Note that in contrast nearest neighbor-based algorithms, K-LPE, aK-LPE or BP-$K$-NNG \cite{ref:Manqi2009,ref:Jing2012,ref:knn_2011}, require $O(nd)$ for testing one point. It is worth noting that $s\leq n$ comes from the ``support pairs'' within the input preference pair set. 
Practically we observe that for most data sets $s$ is much smaller than $n$ in the experiment section, leading to significantly reduced test time compared to aK-LPE, as shown in Table.1.
It is worth mentioning that distributed techniques for speeding up computation of $K$-NN distances \cite{ref:Bhaduri11} can be adopted to further reduce test stage time.

\section{Experiments}
\label{sec:exp}
In this section, we carry out point-wise anomaly detection experiments on synthetic and real-world data sets. We compare our ranking-based approach against density-based methods BP-$K$-NNG \cite{ref:knn_2011} and aK-LPE \cite{ref:Jing2012}, and two other state-of-art methods based on random sub-sampling, isolated forest \cite{ref:isolation_forest} (iForest) and massAD \cite{ref:massAD}.
One-class SVM \cite{ref:oc_svm2001} is included as a baseline.
Other methods such as \cite{ref:Ramaswamy2000,ref:Breunig2000} are not included because they are claimed to be outperformed by above approaches.


\subsection{Implementation Details}
In our simulations, the Euclidean distance is used as distance metric for all candidate methods. For one-class SVM the lib-SVM codes \cite{ref:libsvm}  are used. The algorithm parameter and the RBF kernel parameter for one-class SVM are set using the same configuration as in \cite{ref:massAD}.
For iForest and massAD, we use the codes from the websites of the authors, with the same configuration as in \cite{ref:massAD}.
For aK-LPE we use the average $k$-NN distance 
with fixed $k=20$ since this appears to work better than the actual $K$-NN distance of \cite{ref:Manqi2009}. Note that this is also suggested by the convergence analysis in Thm~1  \cite{ref:Jing2012}. For BP-$K$-NNG, the same $k$ is used and other parameters are set according to \cite{ref:knn_2011}.

For our rankAD approach we follow the steps described in Algorithm 1. We first calculate the ranks $R_n(x_i)$ of nominal points according to Eq.(3) based on a$K$-LPE.
We then quantize $R_n(x_i)$ uniformly into $m$=3 levels $r_q(x_i)\in\{1,2,3\}$ and generate pairs $(i, j)\in\mathcal{P}$ whenever $r_q(x_i)>r_q(x_j)$.
We adapt the routine from \cite{ref:ranksvm_chapelle} and extend it to a kernelized version for the learning-to-rank step Eq.(\ref{eq:ranksvm_standard}).
The trained ranker is then adopted in Eq.(4) for test stage prediction.
We point out some implementation details of our approach as follows.
\begin{enumerate}
  \item 
{\it Resampling:} We follow \cite{ref:Jing2012} and adopt the U-statistic based resampling to compute aK-LPE ranks. We randomly split the data into two equal parts and use one part as ``nearest neighbors'' to calculate the ranks (Eq.~\ref{estimate_p})) for the other part and vice versa. Final ranks are averaged over 20 times of resampling.

\item  
{\it Quantization levels \& K-NN} For real experiments with 2000 nominal training points, we fix $k=20$ and $m=3$. These values are based on noting that the detection performance does not degrade significantly with smaller quantization levels for synthetic data. The $k$ parameter in $K$-NN is chosen to be 20 and is based on Theorem~\ref{thm:rank_main_consistency} and results from synthetic experiments (see below).

 \item      
{\it Cross Validation using pairwise disagreement loss:} For the rank-SVM step we use a 4-fold cross validation to choose the parameters $C$ and $\sigma$. We vary $C\in \{0.001,0.003,0.01,\dots,300,1000\}$, and the RBF kernel parameter $\sigma \in \Sigma = \{2^i\tilde{D}_K,\,i=-10,-9,\dots,9,10\}$, where $\tilde{D}_K$ is the average $20$-NN distance over nominal samples.
      The pair-wise disagreement indicator loss is adopted from \cite{ref:RDPS2012} for evaluating rankers on the input pairs:
  \begin{equation*}
    L(f) = \sum_{(i,j)\in \mathcal{P}} \textbf{1}_{\{ f(x_i) < f(x_j) \}} 
  \end{equation*}  
\end{enumerate}


All reported AUC performances are averaged over 5 runs.


\subsection{Synthetic Data sets}

We first apply our method to a Gaussian toy problem, where the nominal density is: \[ f_0 \sim 0.2 \mathcal{N} \left(
\left[ 5; 0 \right],
\left[ 1, 0 ; 0, 9 \right]
 \right)
+
0.8 \mathcal{N} \left(
\left[ -5 ; 0 \right],
\left[ 9 , 0; 0 , 1 \right]
 \right).
\]
Anomaly follows the uniform distribution within $\{(x,y):\,\,-18\leq x\leq 18, -18\leq y\leq 18\}$. The goal here is to understand the impact of different parameters ($k$-NN parameter and quantization level) used by RankAD. %
Fig.2 shows the level curves for the estimated ranks on the test data. As indicated by the asymptotic consistency (Thm.2) and the finite sample analysis (Thm.3), the empirical level curves of rankAD approximate the level sets of the underlying density quite well.
\begin{figure}[htbp]\label{fig:gaussian}
\begin{centering}
\begin{minipage}[t]{.45\textwidth}
\includegraphics[width = 1\textwidth]{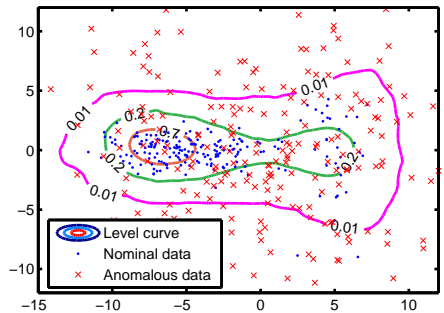}
\end{minipage}
\caption{ Level sets for the estimated ranks. 600 training points are used for training.}
\end{centering}
\end{figure}
%
We vary $k$ and $m$ and evaluate the AUC performances of our approach shown in Table \ref{tab:sensitivity}.
The Bayesian AUC is obtained by thresholding the likelihood ratio using the generative densities.
From Table \ref{tab:sensitivity} we see the detection performance is quite insensitive to the $k$-NN parameter and the quantization level parameter $m$, and for this simple synthetic example is close to Bayesian performance.

\begin{table}[!htbp]
\caption{ AUC performances of Bayesian detector, aK-LPE, and rankAD with different values of $k$ and $m$. 600 training points are used for training. For test 500 nominal and 1000 anomalous points are used. \label{tab:sensitivity} }
\begin{center}
\begin{tabular}{|c||c|c|c|c|}
  \hline
  AUC       & k=5  &  k=10  & k=20 & k=40 \\
  \hline\hline
  m=3       & 0.9206  &  0.9200  &  0.9223  &  0.9210 \\    \hline
  m=5       & 0.9234  &  0.9243  &  0.9247  &  0.9255 \\    \hline
  m=7       & 0.9226  &  0.9228  &  0.9234  &  0.9213 \\    \hline
  m=10      & 0.9201  &  0.9208  &  0.9244  &  0.9196 \\    \hline\hline
  aK-LPE    & 0.9192  &  0.9251  &  0.9244  &  0.9228 \\    \hline
  Bayesian  & 0.9290  &  0.9290  &  0.9290  &  0.9290 \\    \hline
\end{tabular}
\end{center}
\end{table}

\subsection{Real-world data sets}

\begin{table}[!htbp]
\caption{ Data characteristics of the data sets used in experiments. $N$ is the total number of instances. $d$ the dimension of data. The percentage in brackets indicates the percentage of anomalies among total instances. \label{tab:data_sets} }
\begin{center}
\begin{tabular}{|c||c|c|c|}
  \hline
  data sets     & $N$  &  $d$  & anomaly class  \\
  \hline\hline
  Annthyroid     & 6832  &  6  &  classes 1,2 \\    \hline
  Forest Cover   & 286048  &  10  &  class 4 vs. class 2  \\ \hline
  HTTP           & 567497  &  3   &  attack  \\ \hline
  Mamography     & 11183   &  6   &  class 1  \\ \hline
  Mulcross       & 262144  &  4   &  2 clusters  \\ \hline
  Satellite      & 6435    &  36  &  3 smallest classes  \\ \hline
  Shuttle        & 49097   &  9   &  classes 2,3,5,6,7  \\ \hline
  SMTP           & 95156  &  3   &  attack   \\
  \hline
\end{tabular}
\end{center}
\end{table}

We conduct experiments on several real data sets used in \cite{ref:isolation_forest} and \cite{ref:massAD}, including 2 network intrusion data sets HTTP and SMTP from \cite{ref:Yamanishi00}, Annthyroid, Forest Cover Type, Satellite, Shuttle from UCI repository \cite{ref:UCI}, Mammography and Mulcross from \cite{ref:Rocke96}. Table \ref{tab:data_sets} illustrates the characteristics of these data sets.

\begin{table*}[htp]
\caption{Anomaly detection AUC performance and test stage time of various methods.}
\begin{center}
\begin{tabular}{|c|c||c|c|c|c|c|c|}
  \hline
  \multicolumn{2}{|c||}{Data Sets} &   rankAD  &   one-class svm  &   BP-$K$-NNG   & aK-LPE & iForest & massAD \\
  \hline\hline
  \multirow{8}{*}{AUC}
  &    Annthyroid        & 0.844         & 0.681   & 0.823   &  0.753  & {\bf 0.856}   & 0.789 \\
  &   Forest Cover      & {\bf 0.932}   & 0.869   & 0.889   &  0.876  & 0.853         & 0.895 \\
  &    HTTP              & {\bf 0.999}   & 0.998   & 0.995   & {\bf 0.999}  & 0.986         & 0.995 \\
  &    Mamography        & {\bf 0.909}   & 0.863   & 0.886   &  0.879  & 0.891         & 0.701 \\
  &    Mulcross          & {\bf 0.998}   & 0.970   & 0.994   & {\bf 0.998}  & 0.971         & 0.998 \\
  &    Satellite         & {\bf 0.885}   & 0.774   & 0.872   &  0.884  & 0.812         & 0.692 \\
  &    Shuttle           & {\bf 0.996}   & 0.975   & 0.985   &  0.995  & 0.992         & 0.992 \\
  &   SMTP              & {\bf 0.934}   & 0.751   & 0.902   &  0.900  & 0.869         & 0.859 \\
  \hline
  \multirow{8}{*}{test time}
    & Annthyroid        & 0.338  & 0.281 & 2.171	& 2.173  & 1.384  & 0.030 \\
    & Forest Cover      & 1.748  & 1.638 & 8.185	& 13.41  & 7.239  & 0.483 \\
    & HTTP              & 0.187  & 0.376 & 2.391	& 11.04  & 5.657  & 0.384 \\
    & Mamography        & 0.237  & 0.223 & 0.981	& 1.443  & 1.721  & 0.044 \\
    & Mulcross          & 2.732  & 2.272 & 8.772 	& 13.75  & 7.864  & 0.559 \\
    & Satellite         & 0.393  & 0.355 & 0.976 	& 1.199  & 1.435  & 0.030 \\
    & Shuttle           & 1.317  & 1.318 & 6.404 	& 7.169  & 4.301  & 0.186 \\
    & SMTP              & 1.116  & 1.105 & 7.912 	& 11.76  & 5.924  & 0.411 \\
  \hline
\end{tabular}
\end{center}
\label{tab:real_AUC}
\end{table*}

We randomly sample 2000 nominal points for training. The rest of the nominal data and all of the anomalous data are held for testing. Due to memory limit, at most 80000 nominal points are used at test time. The time for testing all test points and the AUC performance are reported in Table \ref{tab:real_AUC}.

We observe that while being faster than BP-$K$-NNG, aK-LPE and iForest, and comparable to one-class SVM during test stage, our approach also achieves superior performance for all data sets.
The density based aK-LPE and BP-$K$-NNG has somewhat good performance, but its test-time degrades with training set size.
massAD is very fast at test stage, but has poor performance for several data sets.

{\it one-class SVM Comparison}
The baseline one-class SVM has good test time due to the similar $O(dS_1)$ test stage complexity where $S_1$ denotes the number of support vectors.
However, its detection performance is pretty poor, because one-class SVM training is in essence approximating one single $\alpha$-percentile density level set. $\alpha$ depends on the parameter of one-class SVM, which essentially controls the fraction of points violating the max-margin constraints \cite{ref:oc_svm2001}.
Decision regions obtained by thresholding with different offsets are simply scaled versions of that particular level set.
Our rankAD approach significantly outperforms one-class SVM, because it has the ability to approximate different density level sets.

{\it aK-LPE \& BP-$K$-NNG Comparison:}
Computationally RankAD significantly outperforms density-based aK-LPE and BP-$K$-NNG, which is not surprising given our discussion in Sec.4.3.
Statistically, RankAD appears to be marginally better than aK-LPE and BP-$K$-NNG for many datasets and this requires more careful reasoning. To evaluate the statistical significance  of the reported test results we note that the number of test samples range from 5000-500000 test samples with at least 500 anomalous points. Consequently, we can bound test-performance to within 2-5\% error with 95\% confidence ($<2\%$ for large datasets and $<5\%$ for the smaller ones (Annthyroid, Mamography, Satellite) ) using standard extension of known results for test-set prediction~\cite{langford05}. After accounting for this confidence RankAD is marginally better than aK-LPE and BP-$K$-NNG statistically. For aK-LPE we use resampling to robustly ranked values (see Sec.~6.1) and for RankAD we use cross-validation (CV) (see Sec.~6.1) for rank prediction. Note that we cannot use CV for tuning predictors for detection because we do not have anomalous data during training. All of these arguments suggests that the regularization step in RankAD results in smoother level sets and better accounts for smoothness of true level sets (also see Fig~\ref{fig:gaussian}) in some cases, unlike NN methods. We plan to investigate this in our future work. 

\section{Conclusions}
\label{sec:con}
In this paper, we propose a novel anomaly detection framework based on RankAD. We combine statistical density information with a discriminative ranking procedure.
Our scheme learns a ranker over all nominal samples based on the $k$-NN distances within the graph constructed from these nominal points. This is achieved through a pair-wise learning-to-rank step, where the inputs are preference pairs $(x_i,x_j)$. The preference relationship for $(x_i,x_j)$ takes a value one if the nearest neighbor based score for $x_i$ is larger than that for $x_j$. Asymptotically this preference models the situation that data point $x_i$ is located in a higher density region relative to $x_j$ under nominal distribution.
We then show the asymptotic consistency of our approach, which allows for flexible false alarm control during test stage.
We also provide a finite-sample generalization bound on the empirical false alarm rate of our approach.
Experiments on synthetic and real data sets demonstrate our approach has state-of-art statistical performance as well as low test time complexity.


\end{document}